\documentclass[10pt,twocolumn,letterpaper]{article}
\pdfoutput=1
 
\usepackage{cvpr}
\usepackage{times}
\usepackage{epsfig}
\usepackage{graphicx}
\usepackage{amsmath}
\usepackage{amssymb}
\usepackage{mathtools,xparse}
\usepackage{diagbox}
\usepackage{verbatim}
\usepackage{setspace}
\usepackage{multirow}
\usepackage[table,xcdraw]{xcolor}

\DeclarePairedDelimiter{\norm}{\lVert}{\rVert}
\newcommand{\HRule}{\rule{0.8\textwidth}{0.2mm}}

\usepackage[pagebackref=true,breaklinks=true,letterpaper=true,colorlinks,bookmarks=false]{hyperref}

\cvprfinalcopy % *** Uncomment this line for the final submission

\def\cvprPaperID{2728} % *** Enter the CVPR Paper ID here

\ifcvprfinal\fi
\begin{document}

%%%%%%%%% TITLE
\title{Gen-LaneNet: A Generalized and Scalable Approach for 3D Lane Detection}

\author{Yuliang Guo\textsuperscript{*} \qquad
Guang Chen \qquad
Peitao Zhao \qquad
Weide Zhang \\
Jinghao Miao \qquad
Jingao Wang \qquad
Tae Eun Choe\\
Baidu Apollo\\
}

\maketitle
%\thispagestyle{empty}

%%%%%%%%% ABSTRACT
\begin{abstract}
We present a generalized and scalable method,  called Gen-LaneNet, to detect 3D lanes from a single image. The method, inspired by the latest state-of-the-art 3D-LaneNet~\cite{Garnett:etal:ICCV2019}, is a unified framework solving image encoding, spatial transform of features and 3D lane prediction in a single network. However, we propose unique designs for Gen-LaneNet in two folds. First, we introduce a new geometry-guided lane anchor representation in a new coordinate frame and apply a specific geometric transformation to directly calculate real 3D lane points from the network output. We demonstrate that aligning the lane points with the underlying top-view features in the new coordinate frame is critical towards a generalized method in handling unfamiliar scenes. Second, we present a scalable two-stage framework that decouples the learning of image segmentation subnetwork and geometry encoding subnetwork. Compared to 3D-LaneNet~\cite{Garnett:etal:ICCV2019}, the proposed Gen-LaneNet drastically reduces the amount of 3D lane labels required to achieve a robust solution in real-world application. Moreover, we release a new synthetic dataset\footnote{\url{https://github.com/yuliangguo/3D_Lane_Synthetic_Dataset}} and its construction strategy to encourage the development and evaluation of 3D lane detection methods. In experiments, we conduct extensive ablation study to substantiate the proposed Gen-LaneNet significantly outperforms 3D-LaneNet~\cite{Garnett:etal:ICCV2019} in average precision(AP) and F-score.
\end{abstract}

\section{Introduction}
\label{sec:intro}

Over the past few years, autonomous driving has drawn numerous attention from both academic and industry. To drive safely, one of the fundamental problems is to perceive the lane structure accurately in real-time. Robust detection on current lane and nearby lanes is not only crucial for lateral vehicle control and accurate localization~\cite{Kogan:etal:ivs2016}, but also a powerful tool to build and validate high definition map~\cite{Homayounfar:etal:CVPR2018}.

\begin{figure}[!h]
    \centering
    \includegraphics[width=0.43\textwidth]{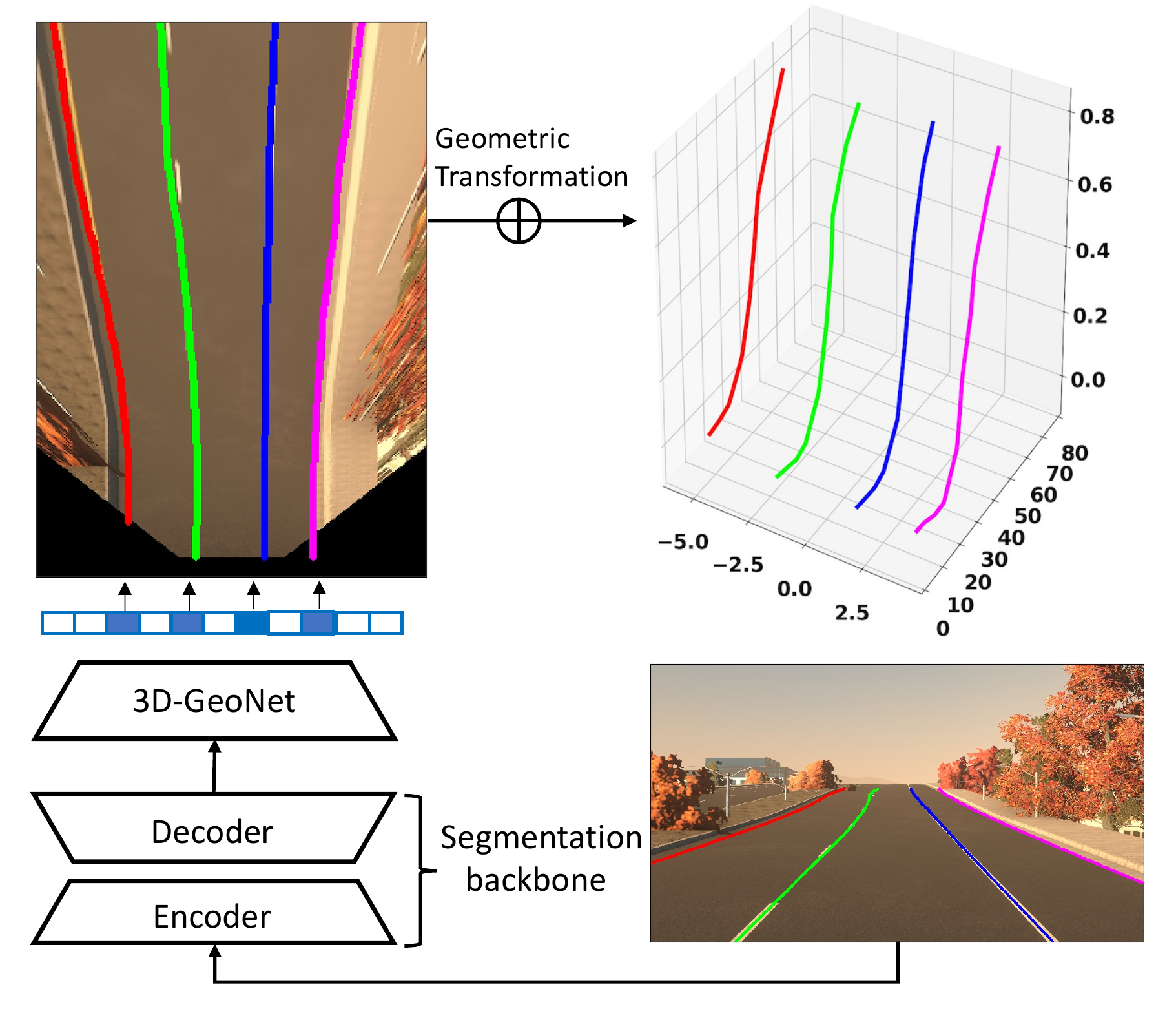}
    \caption{ \small {\bf Procedure of Gen-LaneNet}. A segmentation backbone({\em image segmentation subnetwork}) first encodes an input image in deep features and decodes the features into a lane segmentation map. Given the segmentation as input, 3D-GeoNet({\em geometry encoding subnetwork}) focuses on geometry encoding and predicts intermediate 3D lane points, specifically represented in top-view 2D coordinates and real heights. At last, the presented geometric transformation directly converts the network output to real-world 3D lane points.}
  \label{fig:our:pipeline}
  \vspace{-6mm}
\end{figure}

The majority of image-based lane detection methods treat lane detection as a 2D task~\cite{Tusimple2018,CityScapes2016,Pan:etal:AAAI2018}. A typical 2D lane detection pipeline consists of three components: A semantic segmentation component, which assigns each pixel in an image with a class label to indicate whether it belongs to a lane or not; a spatial transform component to project image segmentation output to a flat ground plane; and a third component to extract lanes which usually involves lane model fitting with strong assumption,{\it e.g.}, fitting quadratic curves. By assuming the world is flat, a 2D lane represented in the flat ground plane might be an acceptable approximation for a 3D lane in the ego-vehicle coordinate system. However, this assumption could lead to unexpected problems, as well studied in~\cite{Garnett:etal:ICCV2019,MultiSensor2019}. For example, when an autonomous driving vehicle encounters a hilly road, an unexpected driving behavior is likely to occur since the 2D planar geometry provides incorrect perception of the 3D road.

To overcome the shortcomings associated with planar road assumption, the latest trend of methods~\cite{YawPitchRool2002,3DLaneStereo2004,MultiSensor2019,Garnett:etal:ICCV2019} has started to focus on perceiving complex 3D lane structures. Specifically, the latest state-of-the-art 3D-LaneNet~\cite{Garnett:etal:ICCV2019} has introduced an end-to-end framework unifying image encoding, spatial transform between image view and top view, and 3D curve extraction in a single network. 3D-LaneNet shows promising results to detect 3D lanes from a monocular camera. However, representing lane anchors in an inappropriate space makes 3D-LaneNet not generalizable to unobserved scenes, while the end-to-end learned framework makes it highly affected by visual variations.

In this paper, we present \textbf{Gen-LaneNet}, a generalized and scalable method to detect 3D lanes from a single image. We introduce a new design of geometry-guided lane anchor representation in a new coordinate frame and apply a specific geometric transformation to directly calculate real 3D lane points from the network output. %In principle the anchor design is an intuitive extension to the anchors of 3D-LaneNet, yet representing the lane anchor in an appropriate coordinate frame is critical for generalization. 
We demonstrate that aligning the anchor representation with the underlying top-view features is critical to make a method generalizable to  unobserved scenes. Moreover, we present a scalable two-stage framework allowing the independent learning of image segmentation subnetwork and geometry encoding subnetwork, which drastically reduces the amount of 3D labels required for learning. Benefiting from more affordable 2D data, a two-stage framework outperforms end-to-end learned framework when expensive 3D labels are rather limited to certain visual variations. At last, we present a highly realistic synthetic dataset of images with rich visual variation, which would serve the development and evaluation of 3D lane detection. In experiments, we conduct extensive ablation study to substantiate that Gen-LaneNet significantly outperforms prior state-of-the-art~\cite{Garnett:etal:ICCV2019} in AP and F-score, as high as 13\% in some test sets.

\section{Related work}
\label{sec:related:work}

Various techniques have been proposed to tackle the lane detection problem. Driven by the effectiveness of Convolutional Neural Network(CNN), lots of recent progress can be observed in improving the 2D lane detection. Some prior methods focus on improving the accuracy of lane segmentation~\cite{DualCNN2016,EmpiricalEval2015,RandomSample2014,HoughLaneMark2012,RobustLane2018,EToE2018,DeepSemantic2018,Pan:etal:AAAI2018,Hou:etal:ICCV2019} while others try to improve segmentation and curve extraction in a unified network~\cite{DNNTraffic2016,Philion:CVPR2019}. %either by combining CNN-based segmentation network with Recurrent Neural Network(RNN)~\cite{DNNTraffic2016} or by a network architecture simulating local contour tracing~\cite{Philion:CVPR2019}. 
More delicate network architectures are further developed to unify 2D lane detection and the following projection to planar road plane into an end-to-end learned network architecture~\cite{Lee:etal:vpgnet:ICCV2017,Kheyrollahi:2012:ARR,DualCNN2016,Neven:etal:IV2018,Brabandere:etal:CoRR2019}. However as discussed in Section~\ref{sec:intro}, all these 2D lane detectors suffer from the specific planar world assumption. Indeed, even perfect 2D lanes are far from sufficient to imply accurate lane positions in 3D space.

As a better alternative, 3D lane detection assumes no planar road and thus provides more reliable road perception. However, 3D lane detection is more challenging, because 3D information is generally unrecoverable from a single image. Consequently, existing methods are rather limited and usually based on multi-sensor or multi-view camera setups~\cite{YawPitchRool2002,3DLaneStereo2004,MultiSensor2019} rather than monocular camera. ~\cite{MultiSensor2019} takes advantage of both LiDAR and camera sensors to detect lanes in real world. But the high cost and high data sparsity of LiDAR limits its practical usage({\it e.g.,} effective detection range is 48 meters in ~\cite{MultiSensor2019}). ~\cite{YawPitchRool2002,3DLaneStereo2004} apply more affordable stereo cameras to perform the 3D lane detection, but they also suffer from low accuracy of 3D information in the distance.

\begin{figure*}[!h]
  \centering
  \includegraphics[width=0.9\textwidth]{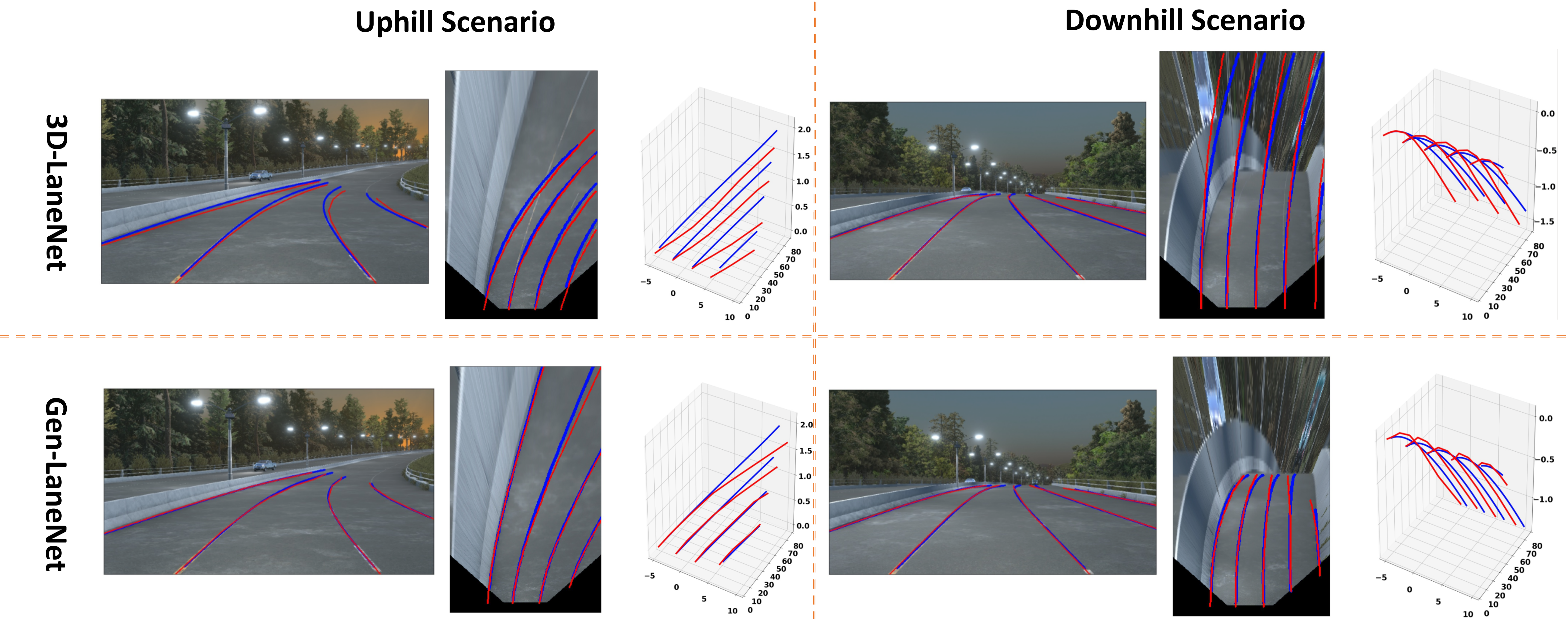}
\caption{{\bf 3D-LaneNet~\cite{Garnett:etal:ICCV2019} and Gen-LaneNet} are compared in two typical scenes with ground height change. We have color-coded ground-truth lanes in blue and predicted lanes in red. Observed from {\bf top-views} in each row, 3D-LaneNet represents anchor points in a coordinate frame not aligned with the underlying visual features(white lane marks). While the proposed Gen-LaneNet resolves this issue.} 
  \label{fig:visual:compare}
\end{figure*}

% \begin{figure}[!h]
%   \centering
%   \includegraphics[width=0.4\textwidth]{figs/images_00_0000285_3DLaneNet_laneline.png}
%   \includegraphics[width=0.4\textwidth]{figs/images_00_0000420_3DLaneNet_laneline.png}\\
%   \includegraphics[width=0.4\textwidth]{figs/images_00_0000285_GenLaneNet_laneline.png}
%   \includegraphics[width=0.4\textwidth]{figs/images_00_0000420_GenLaneNet_laneline.png}
% \caption{{\bf Comparisons between 3D-LaneNet~\cite{Garnett:etal:ICCV2019} and Gen-LaneNet} in two typical scenes with ground height change, uphill and downhill. We have color-coded ground truth lanes in blue and predicted lanes in red. Observed from top-views in the top row, 3D-LaneNet represents anchor points in a coordinate frame not aligned with the underlying visual features. While the proposed Gen-LaneNet resolves this issue.} 
%   \label{fig:visual:compare}
% \end{figure}

% Learning against such ``corrupt'' ground-truth could force a learning toward global memory of the whole scene thus make the model not generalizable to unfamiliar scenes. 

%  by representing anchors in coordinates well-aligned with the visual features, as shown in the bottom row, and applying a specific transform to calculate real 3D lanes afterwards. Moreover, Gen-LaneNet reasons the visibility of lanes from top view and only reserves the visible portions.

The current state-of-the-art, 3D-LaneNet~\cite{Garnett:etal:ICCV2019}, predicts 3D lanes from a single image. It is the first attempt to solve 3D lane detection in a single network unifying image encoding, spatial transform of features and 3D curve extraction. It is realized in an end-to-end learning-based method with a network processing information in two pathways: The \textit{image-view pathway} processes and preserves information from the image while the \textit{top-view pathway} processes features in top-view to output the 3D lane estimations. Image-view pathway features are passed to the top-view pathway through four projective transformation layers which are conceptually built upon the spatial transform network~\cite{Jaderberg:etal:SpatialTransformNet:NIPS2015}. Finally, top-view pathway features are fed into a \textit{lane prediction head} to predict 3D lane points. Specifically, anchor representation of lanes has been developed to enable the \textit{lane prediction head} to estimate 3D lanes in the form of polylines. 3D-LaneNet shows promising results in recovering 3D structure of lanes in frequently observed scenes and common imaging conditions, however, its practicality is questionable due to two major drawbacks.

First, 3D-LaneNet uses an inappropriate coordinate frame in anchor representation, in which ground-truth lanes are misaligned with visual features. This is most evident in the hilly road scenario, where the parallel lanes projected to the virtual top-view appear nonparallel, as observed in the top row of Figure~\ref{fig:visual:compare}. However the ground-truth lanes (blue lines) in 3D coordinate frame are not aligned with the underlying visual features (white lane marks). Training a model against such ``corrupt'' ground-truth could force the model to learn a global encoding of the whole scene. Consequently, the model could hardly generalize to a new scene partially different from an observed one in training.

%%%%%%%%%%%%%%%%%%%%%%%%%%%%%%%%%%%%%%%%%%%%
% move to supplemental if necessary
%%%%%%%%%%%%%%%%%%%%%%%%%%%%%%%%%%%%%%%%%%%%
\begin{comment}
To further demonstrate this fact in feature level, we visualize features from a set of key layers, as shown in Figure~\ref{fig:feat:vis}. Observe that, top-view features produced by~\cite{Garnett:etal:ICCV2019} are indeed not aligned with 3D lane coordinates.

\begin{figure}[!h]
  \centering
%  \includegraphics[width=0.45\textwidth]{figs/feature_visualization.png}
  \includegraphics[height=0.14\textwidth]{figs/feat_exp_img.png}
  \includegraphics[height=0.14\textwidth]{figs/feat_exp_top1.png}
  \includegraphics[height=0.14\textwidth]{figs/feat_exp_top2.png}
  \includegraphics[height=0.14\textwidth]{figs/feat_exp_top3.png}
\caption{\textbf{3D-LaneNet feature visualization:} Given an image captured on a uphill road, imaged lanes are suppose to appear diverging when projected to the virtual top-view. As observed from the visualized features from those blue-marked layers in Figure~\ref{fig:3DLaneNet:Arch}(a), \textbf{top-view features} also form diverging lines. Consequently, diverging visual features will not be in alignment with the parallel lane lines prepared as ground-truth.}
  \label{fig:feat:vis}
\end{figure}
\end{comment}
%%%%%%%%%%%%%%%%%%%%%%%%%%%%%%%%%%%%%%%%%%%%%

Second, the end-to-end learned network indeed makes geometric encoding unavoidably affected by the change of image appearance, because it closely couples 3D geometry reasoning with image encoding. As a result, 3D-LaneNet might require exponentially increased amount of training data in order to reason the same 3D geometry in the presence of partial occlusions, varying illumination or weather conditions. Unfortunately labeling 3D lanes is much more expensive than labeling 2D lanes. It often requires high-definition map built upon expensive multiple sensors (LiDAR, camera, etc), accurate localization and online calibration, and even more expensive manual adjustments in 3D space to produce correct ground truth. These limitations prevent 3D-LaneNet from being scalable in real application. 

\section{Gen-LaneNet}
\label{sec:method}

Motivated by the success of 3D-LaneNet~\cite{Garnett:etal:ICCV2019} and its drawbacks discussed in Section~\ref{sec:related:work}, we propose Gen-LaneNet, a generalized and scalable framework for 3D lane detection. Compared to 3D-LaneNet, Gen-LaneNet is still a unified framework that solves image encoding, spatial transform of features, and 3D curve extraction in a single network. But it involves major differences in two folds: a geometric extension to lane anchor design and a scalable two-stage network that decouples the learning of image encoding and 3D geometry reasoning.

\subsection{Geometry in 3D Lane Detection}
\label{sec:sub:geometry}

\begin{figure}[!h]
  \centering
  \includegraphics[width=0.4\textwidth]{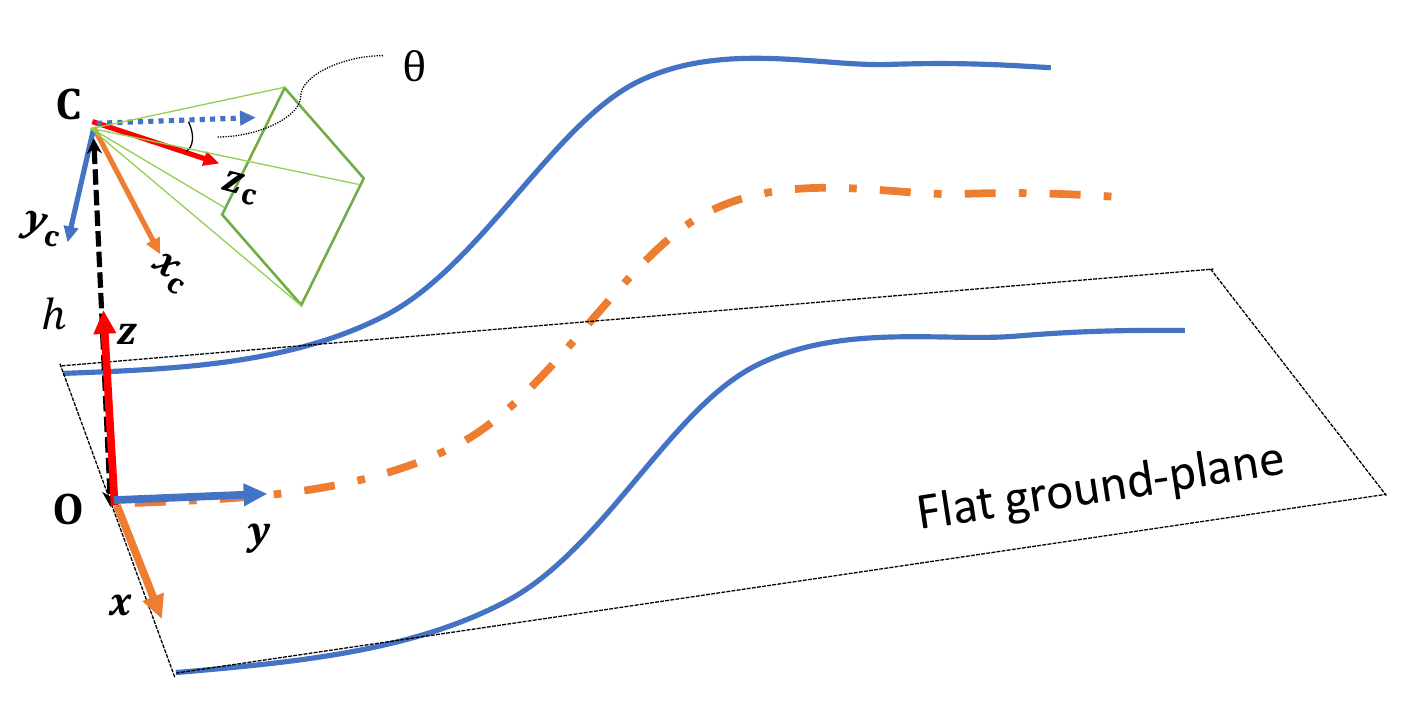}
\caption{\textbf{Camera setup and ego-vehicle coordinate frame.}} %3D lanes are represented in the ego-vehicle coordinate frame defined by $\boldsymbol{x}, \boldsymbol{y}, \boldsymbol{z}$ axes and origin $\boldsymbol{O}$. Specifically, $\boldsymbol{y}$ defines forward, $\boldsymbol{z}$ defines upward and $\boldsymbol{O}$ defines the perpendicular projection of camera center on the road.} %3D lanes are represented in the ego-vehicle coordinate frame defined by $\boldsymbol{x}, \boldsymbol{y}, \boldsymbol{z}$ axes and origin $\boldsymbol{O}$. Specifically, $\boldsymbol{y}$ defines forward, $\boldsymbol{z}$ defines upward and $\boldsymbol{O}$ defines the perpendicular projection of camera center on the road. Following a simple setup, only camera height $h$ and pitch angle $\theta$ are considered to represent camera pose which leads to camera coordinate frame defined by $\boldsymbol{x_c}, \boldsymbol{y_c}, \boldsymbol{z_c}$ axes and origin $\boldsymbol{C}$. A virtual top view can be generated by first projecting a 3D scene to the image plane and then projecting the captured image to the flat ground plane. In this paper, the virtual top-view is considered as a specific 3D coordinate frame defined by $\boldsymbol{\bar{x}}, \boldsymbol{\bar{y}}, \boldsymbol{z}$ axes and origin $\boldsymbol{O}$.}
  \label{fig:coord:sys}
\end{figure}
%%%%%%%%%%%%%%%%%%%%%%%%%%%%%%%%%%%%%%%%%%%%%%%%%%%%%%%%%%%%%%%%%%%%%%%%%%%%%%%%%%

We begin by reviewing the geometry to establish the theory motivating our method. In a common vehicle camera setup as illustrated in Figure~\ref{fig:coord:sys},
3D lanes are represented in the ego-vehicle coordinate frame defined by $\boldsymbol{x}, \boldsymbol{y}, \boldsymbol{z}$ axes and origin $\boldsymbol{O}$. Specifically $\boldsymbol{O}$ defines the perpendicular projection of camera center on the road. Following a simple setup, only camera height $h$ and pitch angle $\theta$ are considered to represent camera pose which leads to camera coordinate frame defined by $\boldsymbol{x_c}, \boldsymbol{y_c}, \boldsymbol{z_c}$ axes and origin $\boldsymbol{C}$. 
A \textit{virtual top-view} can be generated by first projecting a 3D scene to the image plane through a projective transformation and then projecting the captured image to the flat road-plane via a planer homography. Because camera parameters are involved, points in the virtual top-view in principle have different x, y values compared to their corresponding 3D points in the ego-vehicle system. In this paper, we formally considers the virtual top-view as a unique coordinate frame defined by axes $\boldsymbol{\bar{x}}, \boldsymbol{\bar{y}}, \boldsymbol{z}$ and original \textbf{O}. The geometric transformation between virtual top-view coordinate frame and ego-vehicle coordinate frame is derived next.

\begin{figure}[!h]
  \centering
  \includegraphics[height=0.26\textwidth]{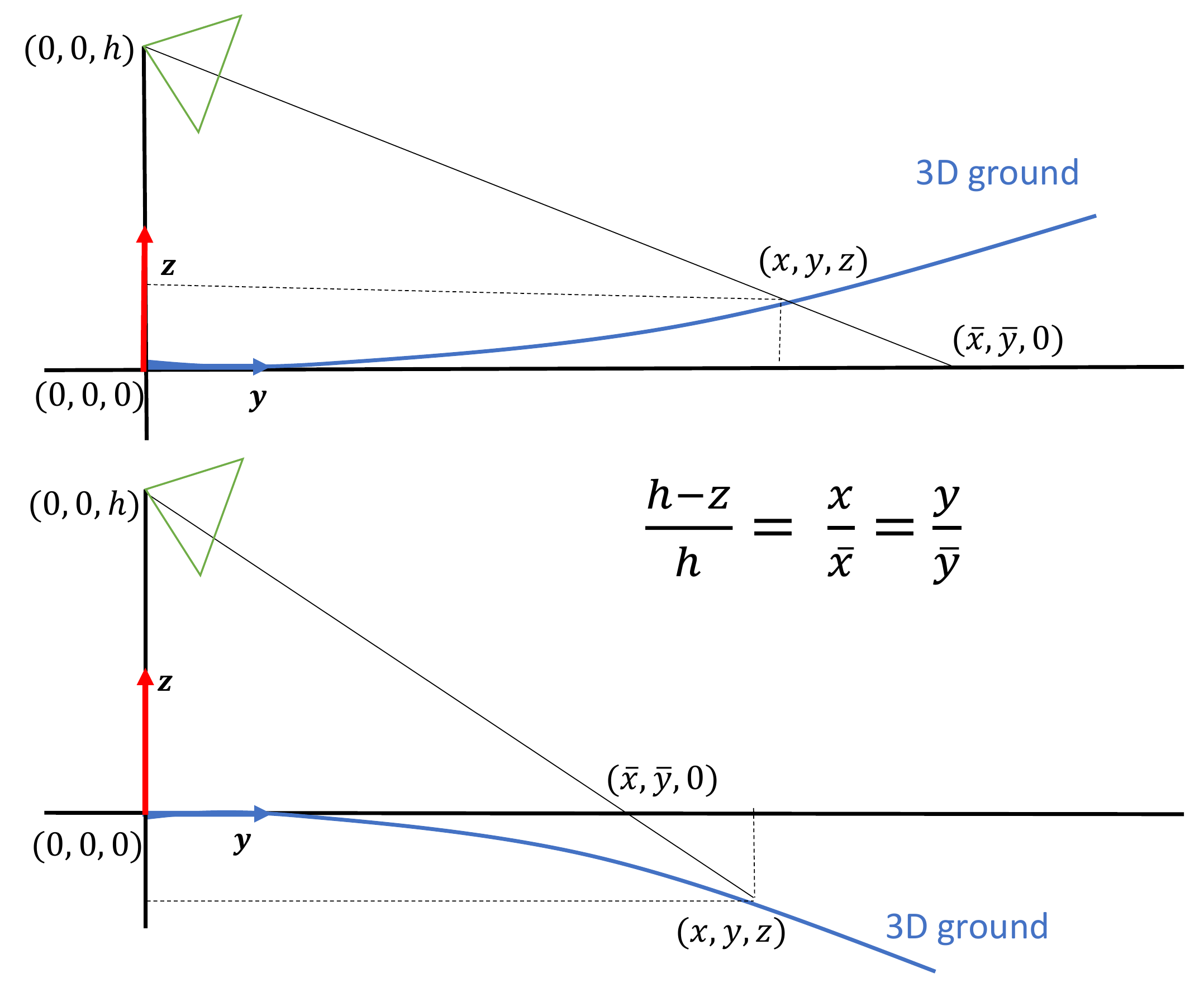}
  \includegraphics[height=0.26\textwidth]{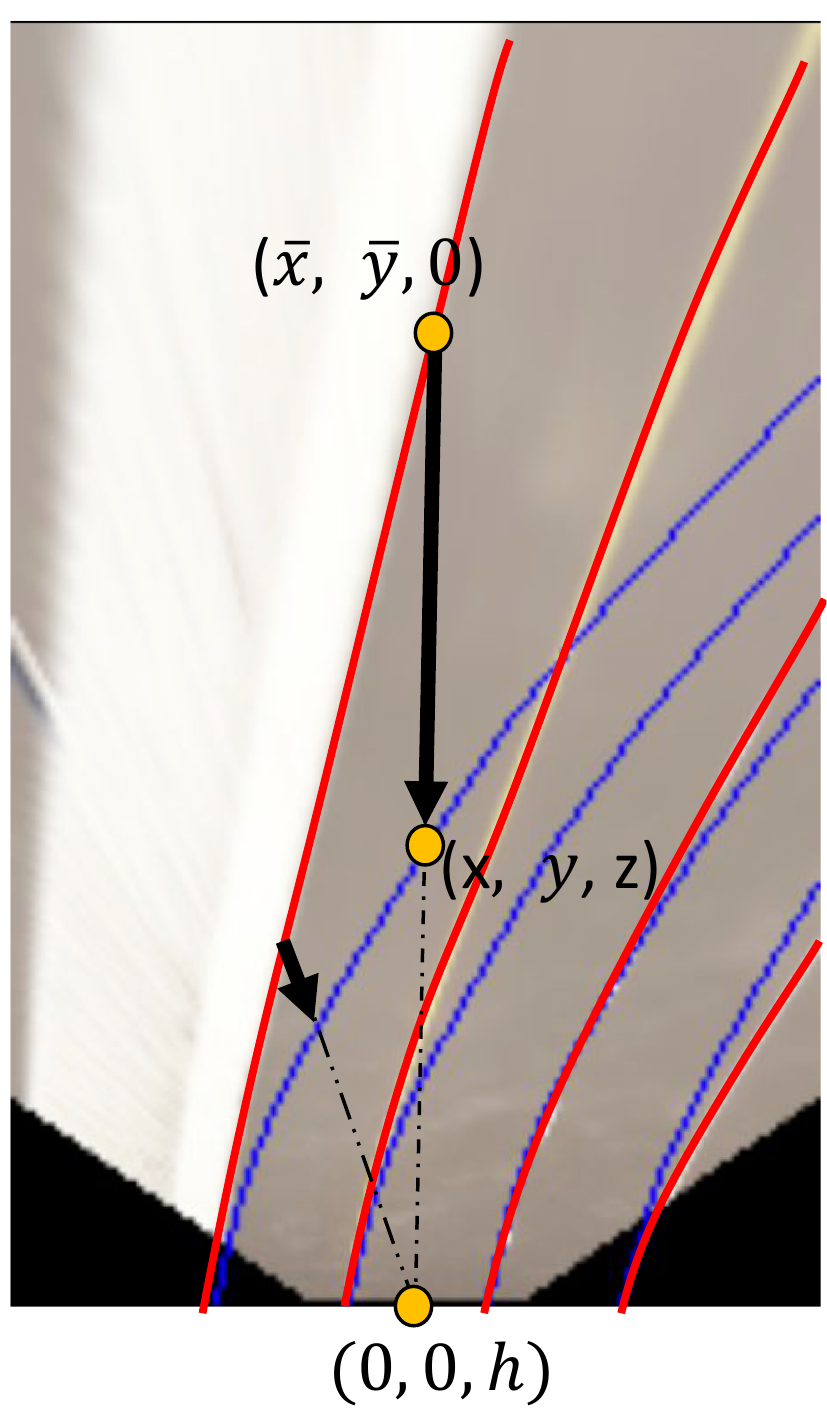}\\
  \hspace{40pt}(a)  \hspace{100pt}(b)
\caption{\textbf{Geometry in 3D lane detection.} (a) The co-linear relationship between a 3D lane point $(x, y, z)$, its projection on the virtual top-view $(\bar{x}, \bar{y}, 0)$ and camera center $(0, 0, h)$ holds, no matter $z>0$ (top) or $z<0$ (bottom). (b) In the virtual top-view, estimating lane height $z$ is conceptually equivalent to estimating the vector field(black arrows) moving top-view lane points (red curves) to their destination positions such that they can form parallel curves (blue curves).}
  \label{fig:geo:transform}
\end{figure}

For a projective camera, a 3D point $(x, y, z)$, its projection on the image plane, and the camera optical center $(0, 0, h)$ should lie on a single ray. Similarly, if a point $(\bar{x}, \bar{y}, 0)$ from the virtual top-view is projected to the same image pixel, it must be on the same ray. Accordingly, camera center $(0, 0, h)$, a 3D point $(x, y, z)$ and its corresponding virtual top-view point $(\bar{x}, \bar{y}, 0)$ appear to be co-linear, as shown in Figure~\ref{fig:geo:transform} (a) and (b). Formally, the relationship between these three points can be written as:

 %The involved spatial transform of features is based on the Inverse Perspective Mapping (IPM), a planar homography calculated from camera intrinsic parameters $K$, and its translation and orientation from the origin of ego-vehicle coordinate frame~\cite{Hartley:2003:MVG}.
%  In a common vehicle camera setup as illustrated in Figure~\ref{fig:coord:sys}, given known $K$, zero roll and yaw angles in camera orientation, the planar homography is determined by only by camera pitch angle $\theta$ and $h$. These two parameters are predicated from the intra-network called\textit{ road plane prediction branch}, as in Figure~\ref{fig:3DLaneNet:Arch} (a). 

\begin{equation}
    \frac{h-z}{h} = \frac{x}{\bar{x}} = \frac{y}{\bar{y}}.
\end{equation}
Specifically, as illustrated in Figure~\ref{fig:geo:transform} (a), this relationship holds no matter $z$ is positive or negative. Thus we derive the geometric transformation from virtual top-view coordinate frame to 3D ego-vehicle coordinate frame as:
\begin{eqnarray}
    \label{eqn:sim_geo}
        x &=& \bar{x} \cdot (1 - \frac{z}{h})\nonumber\\ 
        y &=& \bar{y} \cdot (1 - \frac{z}{h}),
\end{eqnarray}
It is worth mentioning that the obtained transformation describes a general relationship without assuming zero yaw and roll angles in camera orientation. %Although the geometric proof is simply, the algebraic derivation involves more steps. Due to limited space, the detailed derivation is included in Appendix~\ref{sec:algebraic:derive} for interested readers.

\subsection{Geometry-guided anchor representation}
\label{sec:sub:geo_anchor}

\begin{figure*}[!h]
  \centering
  \includegraphics[width=0.95\textwidth]{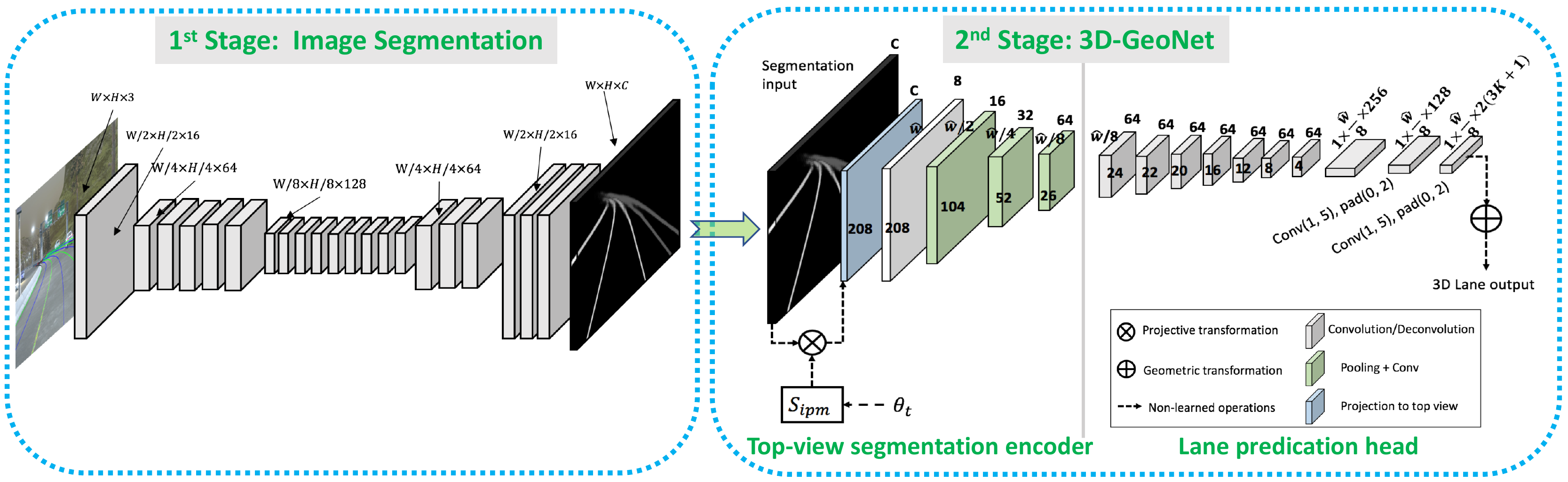}
  
\caption{\textbf{Proposed two-stage network architecture}. Input image with a size $W \times H$ is fed into image segmentation subnetwork in the first stage to generate lane segmentation with the same resolution. The intermediate segmentation map goes through 3D-GeoNet, which is composed of top-view segmentation encoder and lane prediction head, to output 3D lanes represented in virtual top-view. At last, the presented geometric transformation is applied to calculate 3D lane points in ego-vehicle system.}
  \label{fig:GenLaneNet}
\end{figure*}

Following the presented geometry, we solve 3D lane detection in two steps: A network is first applied to encode the image, transform the features to the virtual top-view, and predict lane points represented in virtual top-view; afterwards the presented geometric transformation is adopted to calculate 3D lane points in ego-vehicle coordinate frame, as shown in Figure~\ref{fig:GenLaneNet}. Equation~\ref{eqn:sim_geo} in principle guarantees the feasibility of this approach because the geometric transformation is shown to be independent from camera angles. This is an important fact to ensures the approach not affected by the camera pose estimation.

\begin{figure}[!h]
  \centering
    \includegraphics[width=0.47\textwidth]{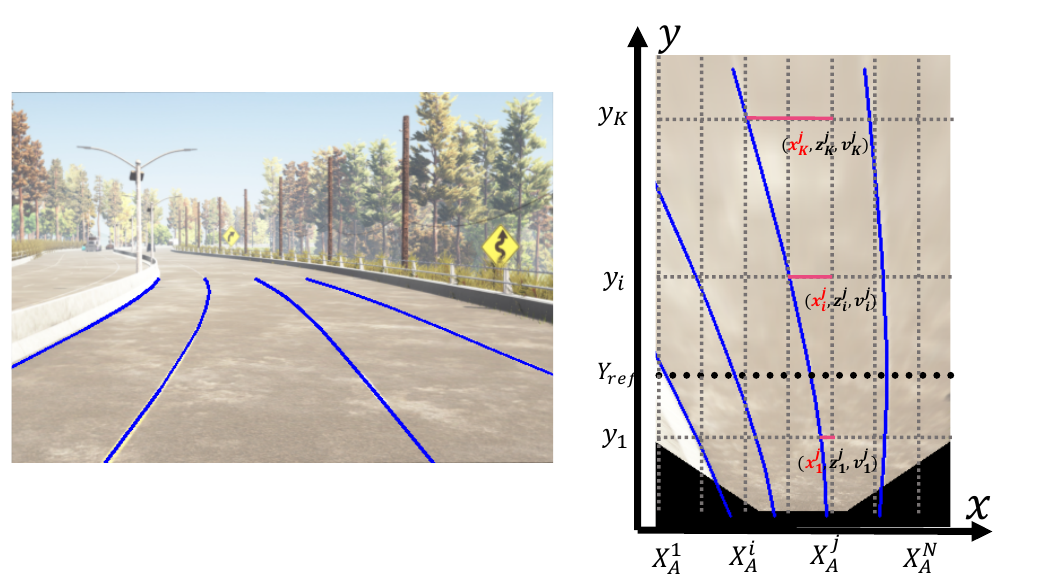}
\caption{\textbf{Anchor representation.} Lane anchors are defined as $N$ equally spaced vertical lines in x-positions $\{X^i_A \}^N_{i=1}$. Given a set of pre-defined fixed y-positions $\{y_i\}^K_{j=1}$, a 3D lane can be represented with an anchor $X^i_A$ composed of  $3 \cdot K$ attributes $\{(\bar{x}^i_j, z^i_j, v^i_j)\}^K_{j=1}$. A ground-truth lane is associated with its closest anchor based on x-value at $Y_{ref}$.}
  \label{fig:Anchors}
\end{figure}

Similar to 3D-LaneNet~\cite{Garnett:etal:ICCV2019}, we develop an anchor representation such that a network can directly predict 3D lanes in the form of polylines. Anchor representation is in fact the essence of network realization of boundary detection and contour grouping in a structured scene. Formally, as shown in Figure~\ref{fig:Anchors}, lane anchors are defined as $N$ equally spaced vertical lines in x-positions $\{X^i_A \}^N_{i=1}$. Given a set of pre-defined fixed y-positions $\{y_i\}^K_{j=1}$, each anchor $X^i_A$ defines a 3D lane line in $3 \cdot K$ attributes $\{(\bar{x}^i_j, z^i_j, v^i_j)\}^K_{j=1}$ or equivalently in three vectors as $(\mathbf{x}^i, \mathbf{z}^i, \mathbf{v}^i)$, where the values $\bar{x}^i_j$ are horizontal offsets relative to the anchor position an the attribute $v^i_j$ indicates the visibility of every lane point. Denoting lane center-line type with $c$ and lane-line type with $l$, each anchor can be written as $X^i_A = \{(\mathbf{x}^i_t, \mathbf{z}^i_t, \mathbf{v}^i_t, p^i_t)\}_{t\in\{c, l\}}$, where $p^t_t$ indicates the existence probability of a lane. Based on this anchor representation, our network outputs 3D lane lines in the virtual top-view. The derived transformation is applied afterwards to calculate their corresponding 3D lane points. Given predicted visibility probability per lane point, only those visible lane points will be kept in the final output.

Our anchor representation involves two major extensions compared to 3D-LaneNet. We represent lane point positions in a different space, the virtual top-view. Representing lane points in the virtual top-view guarantees the target lane position to align with image features projected to the top-view, as shown in the bottom row of Figure~\ref{fig:visual:compare}. Compared with the global encoding of the whole scene in 3D-LaneNet, encoding the correlation at local patch-level is more robust in handling novel or unobserved scenes. Suppose a new scene's overall structure has not been observed from the training, those local patches more likely have. Moreover, we add additional attributes into the anchor representation, to indicate the visibility of each anchor point. As a result, our method is more stable in handling partially visible lanes starting or ending in halfway, as observed in Figure~\ref{fig:visual:compare}.

% However our anchors are defined in virtual top-view coordinate frame, to well aligned with the virtual top-view feature.

% For a 3D lane point, we project its corresponding 2D point in image plane to virtual top-view plane. So the prepared ground truth would be well aligned with virtual top-view feature.

% In data preparation, a predefined $Y_{ref}$ is used to associate a ground-truth 3D lane with the closest anchor $X^i_A$. In order to handle regular lanes, as well as the lane splitting and merging cases, each anchor outputs three types of lane descriptors, two related to lane centerlines and the rest one related to lane delimiter. Throughout a series of convolutions with no padding in y dimension, the lane prediction head reduces feature maps, and final outputs a $N \times 3 \cdot(2 \cdot K + 1)$ dimensional prediction vector.

\subsection{Two-stage framework with decoupled learning of image encoding and geometry reasoning}
\label{sec:two:stage}

Instead of adopting an end-to-end learned network, we propose a two-stage framework which decouples the learning of image encoding and 3D geometry reasoning. As shown in Figure~\ref{fig:GenLaneNet}, the first subnetwork focuses on lane segmentation in image domain; the second predicts 3D lane structure from the segmentation outputs of the first subnetwork. The two-stage framework is well motivated by an important fact that encoding of 3D geometry is rather independent from image features. As observed from Figure~\ref{fig:geo:transform} (a), ground height $z$ is closely correlated to the displacement vector from the position $(\bar{x}, \bar{y})$ to position $(x, y)$. Indeed estimating ground height is conceptually equivalent to estimating a vector field such that all the points corresponding to lanes in the top-view are moved to positions overall in parallelism. When we sort to a network to predict ground height, the network is expected to encode the correlation between visual features and the target vector field. As the target vector field is mostly related to geometry, simple features extracted from a sparse lane segmentation should suffice. 

There are  a  bunch  of  off-the-shelves   candidates~\cite{Romera:etal:ERFNet:TITS2018,ContextNet:2018,Pan:etal:AAAI2018,Hou:etal:ICCV2019} to perform 2D lane segmentation in image, any of which could be effortlessly applied to the first stage in our framework. Although~\cite{ContextNet:2018,Pan:etal:AAAI2018} report better benchmarks performance, we still choose ERFNet~\cite{Romera:etal:ERFNet:TITS2018} for simplicity hence to emphasize the robustness of our framework. For the 3D lane prediction, we propose {\bf 3D-GeoNet}, as in Figure~\ref{fig:GenLaneNet} to estimate 3D lanes from image segmentation. The \textit{top-view segmentation encoder} first projects the segmentation input to a top-view layer and then encodes it in a feature map through a series of convolutional layers. Given the feature map, \textit{lane prediction head} predicts 3D lane attributes based on anchor representation. Upon our anchor representation, lane points produced by \textit{lane prediction head} are represented in top-view positions. 3D lane points in ego-vehicle coordinate frame are calculated afterwards through geometric transformation.

Decoupling the learning of image encoding and geometry reasoning makes the two-stage framework a low-cost and scalable method in real world. As discussed in Section~\ref{sec:related:work}, an end-to-end learned framework like~\cite{Garnett:etal:ICCV2019}, is closely keen to image appearance. Consequently, it depends on huge amount of very expensive real-world 3D data for the leaning. On contrary, the two-stage pipeline drastically reduces the cost as it no longer requires to collect redundant real 3D lane labels in the same area under different weathers, day times, and occlusion cases. Moreover, the two-stage framework could leverage on more sufficient 2D real data, {\it e.g.,} ~\cite{CityScapes2016,Tusimple2018,Pan:etal:AAAI2018} to train a more reliable 2D lane segmentation subnetwork. With extremely robust segmentation as input, 3D lane prediction would in turn perform better. In an optimal situation, the two-stage framework could train the image segmentation subnetwork from 2D real data and train the 3D geometry subnetwork with only synthetic 3D data. We postpone the optimal solution as future work because domain transfer technique is required to resolve the domain gap between perfect synthetic segmentation ground truth and segmentation output from the first subnetwork.
%Consequentially, it improves the algorithm's robustness to potential visual gap between training data and inference data. 

% the pain in collecting real 3D lane labels. The separation of 2D image model learning and geometry model learning provides the feasibility in achieving the goal in theory. 

%An spatial projection and a serios of convolutional layers are applied to transform the images features to top-view and encode in deep features.

\subsection{Training}

Given an image and its corresponding ground-truth 3D lanes, the training proceeds as follows. Each ground-truth lane curve is projected to the virtual top-view, and is associated with the closest anchor at $Y_{ref}$. The ground-truth anchor attributes are calculated based on the ground-truth values at the pre-defined y-positions $\{y_i\}^K_{j=1}$. Given pairs of predicted anchor $X^i_A$ and corresponding ground-truth $\hat{X}^i_A = \{(\mathbf{\hat{x}}^i_t, \mathbf{\hat{z}}^i_t, \mathbf{\hat{v}}^i_t, \hat{p}^i_t)\}_{t\in\{c, l\}}$, the loss function can be written as:
{\small
\begin{flalign}
    \ell = -& \displaystyle\sum_{t\in\{c, l\}} \sum^N_{i=1} (\hat{p}^i_t \log p^i_t + (1- \hat{p}^i_t) \log (1 - p^i_t)) \nonumber\\
           +& \displaystyle\sum_{t\in\{c, l\}} \sum^N_{i=1} \hat{p}^i_t \cdot ( \norm{\mathbf{\hat{v}}^i_t \cdot (\mathbf{x}^i_t - \mathbf{\hat{x}}^i_t)}_1 + \norm{\mathbf{\hat{v}}^i_t \cdot (\mathbf{z}^i_t - \mathbf{\hat{z}}^i_t)}_1) \nonumber \\
           +& \displaystyle\sum_{t\in\{c, l\}} \sum^N_{i=1} \hat{p}^i_t \cdot \norm{\mathbf{v}^i_t - \mathbf{\hat{v}}^i_t}_1
\end{flalign}
}

There are three changes compared to the loss function introduced in 3D-LaneNet~\cite{Garnett:etal:ICCV2019}. First, both $\mathbf{x}^i_t$ and $\mathbf{\hat{x}}^i_t$ are represented in virtual top-view coordinate frame rather than the ego-vehicle coordinate frame. Second, additional cost terms are added to measure the difference between predicted visibility vector and ground-truth visibility vector. Third, cost terms measuring $\bar{x}$ and $z$ distances are multiplied by its corresponding visibility probability $v$ such that those invisible points do not contribute to the loss. %Otherwise, miscalculated loss may jeopardize the model optimization.

%%%%%%%%%%%%%%%%%%%%%%%%%%%%%%%%%%%%%%%%%%%%%%%%%%%%%%%%%%%%%%%%%%%%%%%%%%%%%%%%%%%%%%%%%%%%%%
%%%%%%%%%%%%%%%%%%%%%%%%%%%%%%%%%%%%%%%%%%%%%%%%%%%%%%%%%%%%%%%%%%%%%%%%%%%%%%%%%%%%%%%%%%%%%%
%%%%%%%%%%%%%%%%%%%%%%%%%%%%%%%%%%%%%%%%%%%%%%%%%%%%%%%%%%%%%%%%%%%%%%%%%%%%%%%%%%%%%%%%%%%%%%
%%%%%%%%%%%%%%%%%%%%%%%%%%%%%%%%%%%%%%%%%%%%%%%%%%%%%%%%%%%%%%%%%%%%%%%%%%%%%%%%%%%%%%%%%%%%%%
%%%%%%%%%%%%%%%%%%%%%%%%%%%%%%%%%%%%%%%%%%%%%%%%%%%%%%%%%%%%%%%%%%%%%%%%%%%%%%%%%%%%%%%%%%%%%%

\begin{figure*}[!h]
  \centering
  \includegraphics[width=0.25\textwidth]{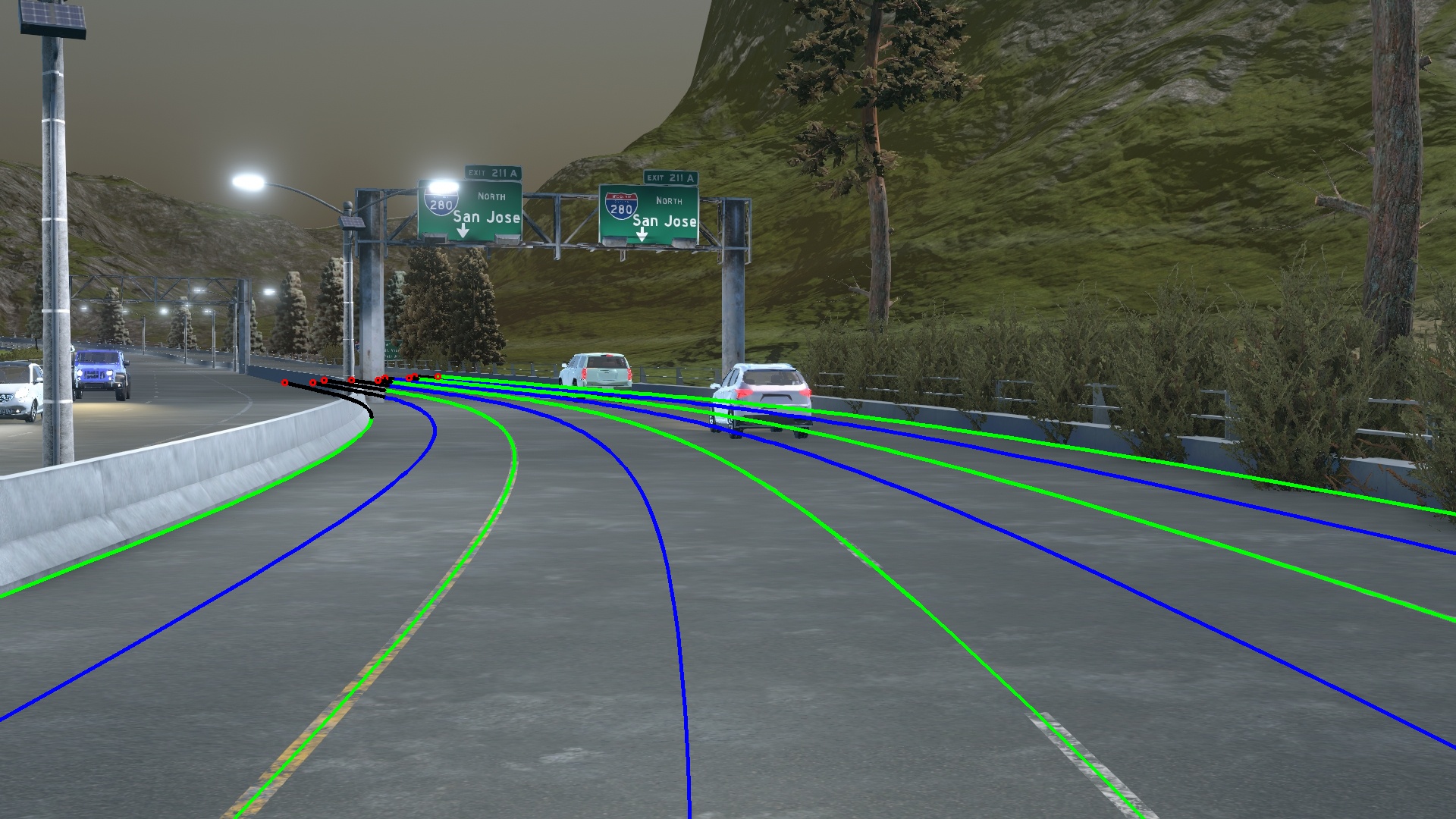}
  \includegraphics[width=0.25\textwidth]{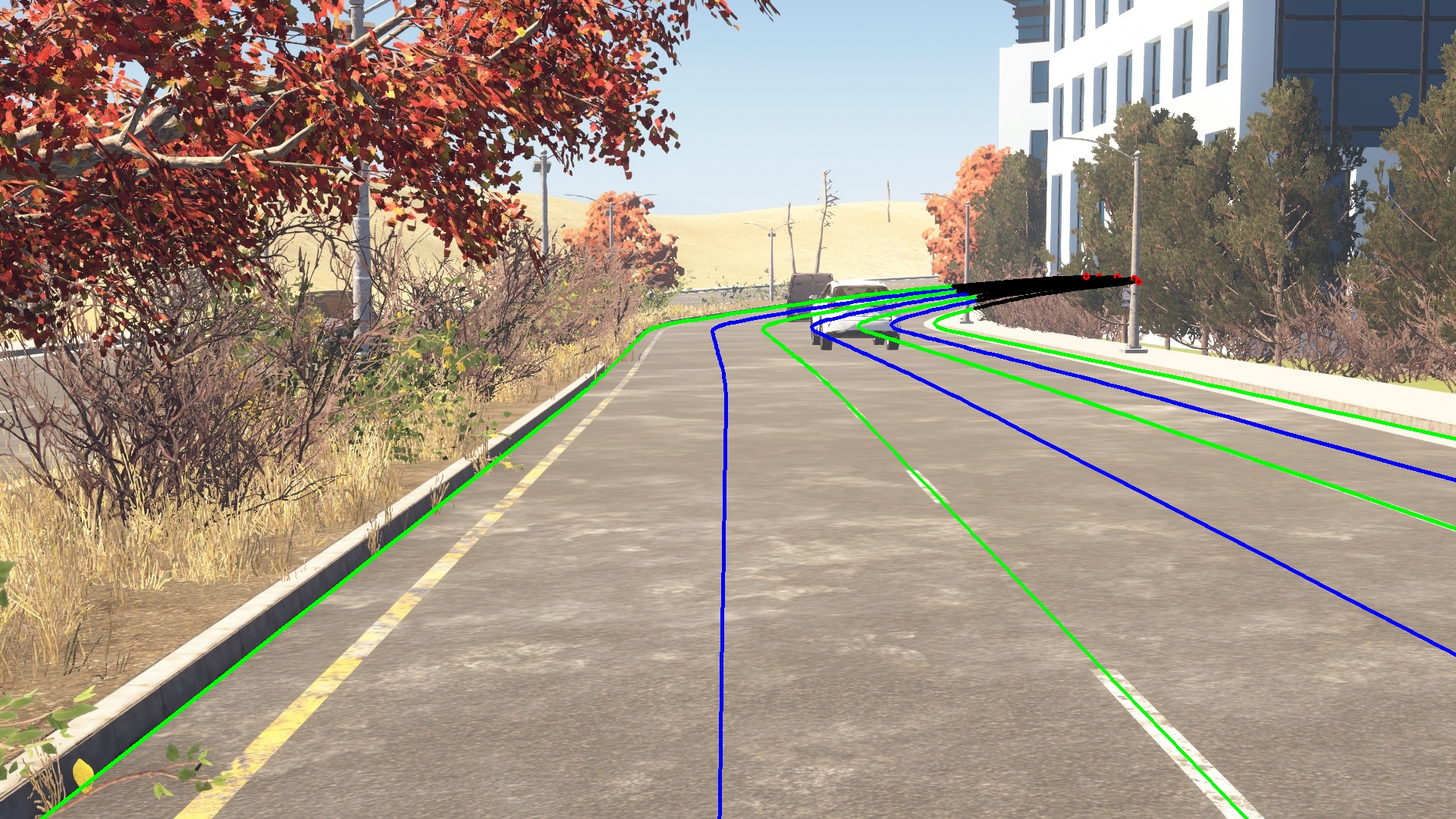}
  \includegraphics[width=0.25\textwidth]{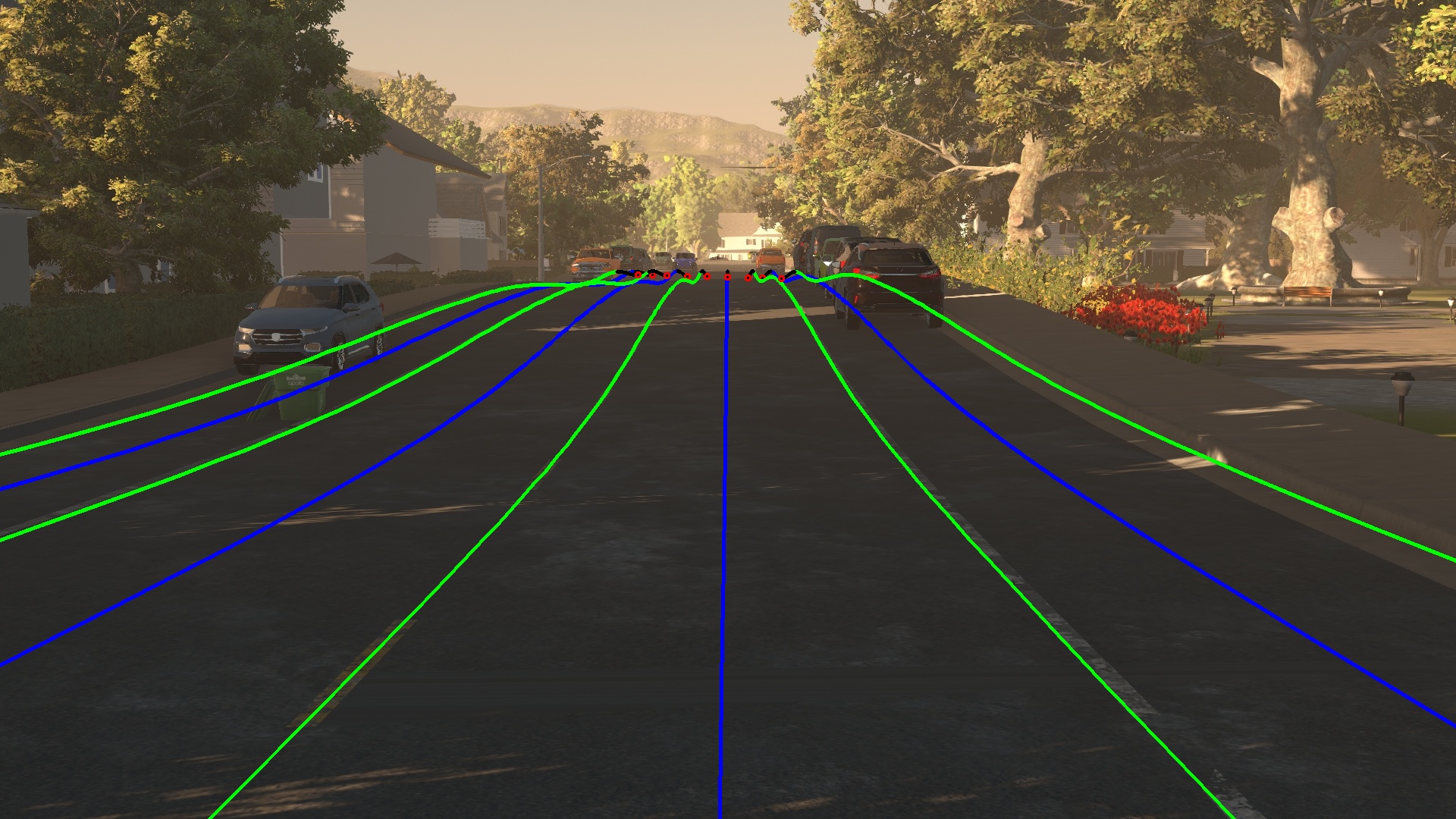}
\caption{\textbf{Examples of synthetic data.} From left to right, images are rendered from  highway map, urban map, and residential map with different day-times respectively. In each image, lane lines and center lines are drawn in green and blue separately. Those black-colored segments of lanes in the distance are discarded in a post-process, as background-occluded segments are generally not desired from a lane detection method.}
  \label{fig:simu:example}
\end{figure*}

\begin{comment}
\begin{figure}[!h]
  \centering
  \includegraphics[width=0.23\textwidth]{figs/00_0000045.jpg}
  \includegraphics[width=0.23\textwidth]{figs/08_0000003.jpg}
  %\includegraphics[width=0.3\textwidth]{figs/16_0000077.jpg}
\caption{\textbf{Examples of synthetic data.} Images are rendered from different terrain maps and with different day-times respectively. In each image, lane lines and center lines are drawn in green and blue separately. Those background-occluded segments of lanes in the distance (black) are discarded in a post-process.}
  \label{fig:simu:example}
\end{figure}
\end{comment}

\section{Synthetic dataset and construction strategy}
\label{sec:data}

Due to lack of 3D lane detection benchmark, we have built a synthetic dataset to develop and validate 3D lane detection methods. Our dataset simulates abundant visual elements and specifically focuses on evaluating a method's generalization capability to rarely observed scenarios. We use Unity game engine to build highly diverse 3D worlds with realistic background elements and render images with diversified scene structure and visual appearance.

The synthetic dataset is rendered from three world maps with diverse terrain information: a highway area, an urban area and a residential area. All the maps are based on real regions at the Silicon Valley in the United States, where lane lines and center lines involve adequate ground height variation and turnings, as shown in Figure~\ref{fig:simu:example}. Images are sparsely rendered at different locations and different day-times(morning, noon, evening), with two levels of lane-marker degradation, random camera-height within $1.4\sim1.8 m$ and random pitch angles within $0^\circ \sim  10^\circ$. We have fixed intrinsic parameters during data rendering and placed a decent amount of agent vehicles driving in the simulation environment, such that the rendered image includes realistic occlusions of lanes. In summary, a total of 6000 samples from virtual high-way map, 1500 samples from urban map, and 3000 samples from residential area, along with corresponding depth map, semantic segmentation map, and 3D lane lines information are provided. 3D lane labels are truncated at 200 meters distance to the camera, and at the border of the rendered image.

So far, essential information about occlusion is still missing for developing reliable 3D lane detectors. In general, a lane detector is expected to recover the foreground-occluded portion but discard the background-occluded portion of lanes, which in turn requires accurate labeling of the occlusion type per lane point. In our dataset, we use ground-truth depth maps and semantic segmentation maps to deduce the occlusion type of lane points. First, a lane point is considered occluded when its $y$ position is deviated from the value at the corresponding pixel in depth map. Second, its occlusion type is further determined based on semantic segmentation map. The final dataset keeps the portion of lanes occluded by foreground but discard the portion occluded by background, as the black segments in the distance shown in Figure~\ref{fig:simu:example}.

%%%%%%%%%%%%%%%%%%%%%%%%%%%%%%%%%%%%%%%%%%%%%%%%%%%%%%%%%%%%%%%%%%%%%%%%%%%%%%%%%%%%%%%%%%%%%%
%%%%%%%%%%%%%%%%%%%%%%%%%%%%%%%%%%%%%%%%%%%%%%%%%%%%%%%%%%%%%%%%%%%%%%%%%%%%%%%%%%%%%%%%%%%%%%
%%%%%%%%%%%%%%%%%%%%%%%%%%%%%%%%%%%%%%%%%%%%%%%%%%%%%%%%%%%%%%%%%%%%%%%%%%%%%%%%%%%%%%%%%%%%%%
%%%%%%%%%%%%%%%%%%%%%%%%%%%%%%%%%%%%%%%%%%%%%%%%%%%%%%%%%%%%%%%%%%%%%%%%%%%%%%%%%%%%%%%%%%%%%%
%%%%%%%%%%%%%%%%%%%%%%%%%%%%%%%%%%%%%%%%%%%%%%%%%%%%%%%%%%%%%%%%%%%%%%%%%%%%%%%%%%%%%%%%%%%%%%

\section{Experiments}
\label{sec:exp}

In the section, we first describe the experimental setups, including dataset splits, baselines, algorithm implementation details, and evaluation metrics. Then we conduct experiments to demonstrate our contributions in ablation. Finally, we design and conduct experiments to substantiate the advantages of our method, compared with prior state of the art~\cite{Garnett:etal:ICCV2019}.

  \subsection{Experimental setup}
  % Overall, we consider three different networks. First, the original 3D-LaneNet~\cite{Garnett:etal:ICCV2019} as shown in Figure~\ref{fig:3DLaneNet:Arch}. Second, 3D-GeoNet, which predicts 3D lanes from perfect semantic segmentation input, as shown in Figure~\ref{fig:GenLaneNet}. The last one, Gen-LaneNet as shown in Figure~\ref{fig:our:pipeline}, is a two-stage network combining a image segmentation network and a geometry network namely 3D-GeoNet. 
{\bf Dataset setup:} In order to evaluate algorithms from different perspectives, we design three different rules to split the synthetic dataset:

(1) {\it Balanced scenes:} the training and testing set follow a standard five-fold split of the whole dataset, to benchmark  algorithms with massive, unbiased data.

(2) {\it Rarely observed scenes:} This dataset split contains the same training data as {\it balanced scenes}, but uses only a subset of the testing data, captured from the complex urban map. This dataset split is designed to examine a method's capability of generalization to the test data rarely observed from training. Because the testing images are sparsely rendered at different locations involving drastic elevation change and sharp turnings, the scenes in testing data are rarely observed from the training data.

(3) {\it Scenes with visual variations:} This split of dataset evaluates methods under the change of illumination, assuming more affordable 2D data compared to expensive 3D data is available to cover the illumination change for the same region. Specifically, the same training set as  {\it balanced scenes} is used to train image segmentation subnetwork in the first stage of our Gen-LaneNet. However 3D examples from a certain day time, namely before dawn, are excluded from the training of 3D geometry subnetwork of our method(3D-GeoNet) and 3D-LaneNet~\cite{Garnett:etal:ICCV2019}. In testing, on contrary, only examples corresponding to the excluded day time are used.

\begin{table*}[]
\centering
\begin{tabular}{|c|c|c|c|c|c|c|c|c|c|c|}
\hline
\multicolumn{2}{|c|}{}                                                                                   & \multicolumn{3}{c|}{\textbf{balanced scenes}} & \multicolumn{3}{c|}{\textbf{rarely observed}} & \multicolumn{3}{c|}{\textbf{visual variations}} \\ \cline{3-11} 
\multicolumn{2}{|c|}{\multirow{-2}{*}{\textbf{\begin{tabular}[c]{@{}c@{}}\end{tabular}}}} & w/o                  & w/                   & gain                                       & w/o                    & w/                     & gain                                          & w/o                      & w/                       & gain                                             \\ \hline
                                                               & \textbf{F-score}                        & 86.4                 & 90.0                 & {\color[HTML]{FE0000} +3.6}                & 72.0                   & 80.9                   & {\color[HTML]{FE0000} +8.9}                   & 72.5                     & 82.7                     & {\color[HTML]{FE0000} +10.5}                     \\ \cline{2-11} 
\multirow{-2}{*}{\textbf{3D-LaneNet}}                          & \textbf{AP}                             & 89.3                 & 92.0                 & {\color[HTML]{FE0000} +2.7}                & 74.6                   & 82.0                   & {\color[HTML]{FE0000} +7.4}                   & 74.9                     & 84.8                     & {\color[HTML]{FE0000} +9.9}                      \\ \hline
                                                               & \textbf{F-score}                        & 88.5                 & 91.8                 & {\color[HTML]{FE0000} +3.3}                & 75.4                   & 84.7                   & {\color[HTML]{FE0000} +9.3}                   & 83.8                     & 90.2                     & {\color[HTML]{FE0000} +6.4}                      \\ \cline{2-11} 
\multirow{-2}{*}{\textbf{3D-GeoNet}}                           & \textbf{AP}                             & 91.3                 & 93.8                 & {\color[HTML]{FE0000} +2.5}                & 79.0                   & 86.6                   & {\color[HTML]{FE0000} +7.6}                   & 86.3                     & 92.3                     & {\color[HTML]{FE0000} +6.0}                      \\ \hline
                                                               & \textbf{F-score}                        & 85.1                 & 88.1                 & {\color[HTML]{FE0000} +3.0}                & 70.0                   & 78.0                   & {\color[HTML]{FE0000} +8.0}                   & 80.9                     & 85.3                     & {\color[HTML]{FE0000} +4.4}                      \\ \cline{2-11} 
\multirow{-2}{*}{\textbf{Gen-LaneNet}}                         & \textbf{AP}                             & 87.6                 & 90.1                 & {\color[HTML]{FE0000} +2.5}                & 73.0                   & 79.0                   & {\color[HTML]{FE0000} +6.0}                   & 83.8                     & 87.2                     & {\color[HTML]{FE0000} +3.4}                      \\ \hline
\end{tabular}
\caption{Comparison of anchor representations. "w/o" represents the integration with anchor design in~\cite{Garnett:etal:ICCV2019}, while "w" represents the integration with our anchor design. For convenience, we also shows the performance gain by integrating our anchor design.}
\label{tab:anchor:compare}
\end{table*}

{\bf Baselines and parameters:} Gen-LaneNet is compared to two other methods: Prior state-of-the-art 3D-LaneNet~\cite{Garnett:etal:ICCV2019}\footnote{We re-implement 3D-LaneNet and reproduce its result reported on Tusimple Dataset~\cite{Tusimple2018}.} is considered as a major baseline; To honestly study the upper bound of our two-stage framework, we treats 3D-GeoNet subnetwork as a stand-alone method which is fed with ground-truth 2D lane segmentation. To conduct fair comparison, all the methods resize the original image into size $360\times480$ and use the same spatial resolution $208\times108$ for the first top-view layer to represent a flat-ground region with the range of $[{-10}, 10] \times [1, 101]$ meters along $\boldsymbol{x}$ and $\boldsymbol{y}$ axes respectively. For the anchor representation, we use y-positions $\{3, 5 , 10, 15, 20, 30, 40, 50, 65, 80, 100\}$, where the intervals are gradually increasing due to the fact that visual information in the distance gets sparser in top-view. In label preparation, we set $Y_{ref}=5$ to associate each lane label with its closest anchor. All the experiments are conducted under known camera poses, intrinsic parameters provided from the synthetic dataset. All the networks are randomly initialized with normal distribution and trained from scratch with Adam optimization and with an initial learning rate $5\cdot10^{-4}$. We set batch size $8$ and complete training in $30$ epochs. For training ERFNet, we follows the same procedure described in~\cite{Romera:etal:ERFNet:TITS2018}, but with modified input image size and output segmentation maps sizes.

{\bf Evaluation metrics:} We formulate the evaluation of 3D lane detection as a bipartite matching problem between predicted lanes and ground-truth lanes. The global best matching is sought via minimum-cost flow. Our evaluation method is so far the most strict compared to one-to-many matching in~\cite{Tusimple2018} or greedy search bipartite matching in~\cite{Garnett:etal:ICCV2019}. 

% Our method differentiates from TuSimple lane evaluation~\cite{Tusimple2018} that allows one-to-many match, and also differentiates from the evaluation method from 3D-LaneNet~\cite{Garnett:etal:ICCV2019} which solves bipartite match in a greedy search.
%Instead, after enumerating the pairwise costs between all pairs of lane lines, the perfect bipartite match between two sets can be sought via solving a minimum-cost flow problem.

To handle partial matching properly, we define a new pairwise cost between lanes in euclidean distance. Specifically, lanes are represented in $X^j = \{x^j_i, z^j_i, v^j_i\}_{i=1}^n$ at $n$ pre-determined y-positions, where $v^i$ indicates whether the y-position is covered by a given lane. Denser y-positions compared to the anchor points are used here, which are equally placed from $0$ to $100$ meters with $2$ meter interval. Formally, the lane-to-lane cost between $X^j$ and $X^k$ is calculated as the square root of the squared sum of point-wise distances over all y-positions, written as $cost_{jk} = \sqrt{\sum_i^n d^{jk}_i}$, where
{\small
\begin{equation}
        d^{jk}_i =
        \begin{cases}
            (x^j_i - x^k_i)^2 + (z^j_i - z^k_i)^2,& \text{if } v^j_i = 1 \text{ and } v^k_i = 1\\
            0,              & \text{if } v^j_i = 0 \text{ and } v^k_i = 0\\
            d_{max},            & \text{otherwise. }
        \end{cases} \nonumber
\end{equation}
}
Specifically, point-wise euclidean distance is calculated when a y-position is covered by both lanes. When a y-position is only covered by one lane, the point-wise distance is assigned to a max-allowed distance $d_{max} = 1.5m$. While a y-position is not covered by any of the lanes, the point-wise distance is set to zero. Following such metric, a pair of lanes covering different ranges of y-positions can still be matched, but at an additional cost proportional to the number of \textit{edited} points. This defined cost is inspired by the concept of \textit{edit distance} in string matching. After enumerating all pairwise costs between two sets, we adopt the solver included in Google OR-tools to solve the minimum-cost flow problem. Per lane from each set, we consider it matched when 75\% of its covered y-positions have point-wise distance less than the max-allowed distance (1.5 meters).
% as illustrated in Figure~\ref{fig:illus:eval}.

At last, the percentage of matched ground-truth lanes is reported as recall and the percentage of matched predicted lanes is reported as precision. We report both the Average Precision(AP) as a comprehensive evaluation and the maximum F-score as an evaluation of the best operation point in application.

\begin{comment}
\begin{figure}[!h]
  \centering
  \includegraphics[width=0.47\textwidth]{figs/illus_eval.png}
  \includegraphics[width=0.47\textwidth]{figs/illus_eval_2.jpg}
\caption{\textbf{Visualization of evaluation result.} We show two pairs of results, both in image and in 3D coordinates. The matching results are visualized in distinguished colors: the  matched  lane  lines  from prediction and ground-truth  are drawn in red and blue respectively; The false negative ground-truth lanes are shown in cyan, and false positive predicted lanes are shown in purple.}
  \label{fig:illus:eval}
\end{figure}
\end{comment}

\subsection{Anchor effect}

We first demonstrate the superiority of the presented geometry-guided anchor representation compared to~\cite{Garnett:etal:ICCV2019}. For each candidate method, we keep the architecture exact the same, except the anchor representation integrated. As reported in Table~\ref{tab:anchor:compare}, all the three methods, no matter end-to-end 3D-LaneNet~\cite{Garnett:etal:ICCV2019}, ``theoretical existing" 3D-GeoNet, or our two-stage Gen-LaneNet, benefit significantly from the new anchor design. Both AP and F-score achieve 3\% to 10\% improvements, across all splits of dataset. %The significant improvements demonstrate the superiority of our geometry-guided lane anchor design.

\begin{table*}[]
\centering
\begin{tabular}{|c|c|c|c|c|}
\hline
\multicolumn{2}{|c|}{\textbf{}} & \textbf{balanced  scenes} & \textbf{rarely observed} & \textbf{visual variations} \\ \hline
 & \textbf{F-score} & 86.4 & 72.0 & 72.5 \\ \cline{2-5} 
\multirow{-2}{*}{\textbf{3D-LaneNet}} & \textbf{AP} & 89.3 & 74.6 & 74.9 \\ \hline
 & \textbf{F-score} & {\color[HTML]{3166FF} \textbf{91.8}} & {\color[HTML]{3166FF} \textbf{84.7}} & {\color[HTML]{3166FF} \textbf{90.2}} \\ \cline{2-5} 
\multirow{-2}{*}{\textbf{3D-GeoNet}} & \textbf{AP} & {\color[HTML]{3166FF} \textbf{93.8}} & {\color[HTML]{3166FF} \textbf{86.6}} & {\color[HTML]{3166FF} \textbf{92.3}} \\ \hline
 & \textbf{F-score} & \textbf{88.1} & \textbf{78.0} & \textbf{85.3} \\ \cline{2-5} 
\multirow{-2}{*}{\textbf{Gen-LaneNet}} & \textbf{AP} & \textbf{90.1} & \textbf{79.0} & \textbf{87.2} \\ \hline
\end{tabular}
\caption{Upper bound of the proposed two-stage framework. 3D-GeoNet shows potential improvement on Gen-LaneNet when a better image segmentation algorithm is integrated.}
\label{tab:twostage}
\end{table*}

\begin{table*}[]
\centering
\begin{tabular}{|c|c|c|c|c|c|c|c|}
\hline
\textbf{Dataset Splits} & \textbf{Method} & \textbf{F-Score} & \textbf{AP} & \textbf{\begin{tabular}[c]{@{}c@{}}x error\\ near (m)\end{tabular}} & \textbf{\begin{tabular}[c]{@{}c@{}}x error \\ far (m)\end{tabular}} & \textbf{\begin{tabular}[c]{@{}c@{}}z error\\ near (m)\end{tabular}} & \textbf{\begin{tabular}[c]{@{}c@{}}z error\\ far (m)\end{tabular}} \\ \hline
 & 3D-LaneNet & 86.4 & 89.3 & 0.068 & 0.477 & 0.015 & 0.202 \\ \cline{2-8} 
\multirow{-2}{*}{\textbf{\begin{tabular}[c]{@{}c@{}}balanced \\  scenes\end{tabular}}} & Gen-LaneNet & {\color[HTML]{FE0000} 88.1} & {\color[HTML]{FE0000} 90.1} & 0.061 & 0.496 & 0.012 & 0.214 \\ \hline
 & 3D-LaneNet & 72.0 & 74.6 & 0.166 & 0.855 & 0.039 & 0.521 \\ \cline{2-8} 
\multirow{-2}{*}{\textbf{\begin{tabular}[c]{@{}c@{}}rarely\\ observed\end{tabular}}} & Gen-LaneNet & {\color[HTML]{FE0000} 78.0} & {\color[HTML]{FE0000} 79.0} & 0.139 & 0.903 & 0.030 & 0.539 \\ \hline
 & 3D-LaneNet & 72.5 & 74.9 & 0.115 & 0.601 & 0.032 & 0.230 \\ \cline{2-8} 
\multirow{-2}{*}{\textbf{\begin{tabular}[c]{@{}c@{}}visual\\ variations\end{tabular}}} & Gen-LaneNet & {\color[HTML]{FE0000} 85.3} & {\color[HTML]{FE0000} 87.2} & 0.074 & 0.538 & 0.015 & 0.232 \\ \hline
\end{tabular}
\caption{Whole system comparison between 3D-LaneNet~\cite{Garnett:etal:ICCV2019} and Gen-LaneNet. }
\label{tab:whole}
\end{table*}

\subsection{Upper bound of two-stage framework}

Experiments are designed to substantiate that a two-stage method potentially gains higher accuracy when more robust image segmentation is provided, and meanwhile to localize the upper bound of Gen-LaneNet when perfect image segmentation subnetwork is provided. As shown in Table~\ref{tab:twostage}, 3D-GeoNet consistently outperforms Gen-LaneNet and 3D-LaneNet across all three experimental setups. We notice that on {\it balanced scenes}, the improvement over Gen-LaneNet is pretty obvious, around 3\% better, while on {\it rarely observed scenes} and {\it scenes with visual variations}, the improvement is significant from 5\% to 7\%. This observation is rather encouraging because the 3D geometry from hard cases({\it e.g.,}new scenes or images with dramatic visual variations) can still be reasoned well from the abstract ground-truth segmentation or from the output of image segmentation subnetwork. Besides, Table~\ref{tab:twostage} also shows promising upper bound of our method, as the 3D-GeoNet outperforms 3D-LaneNet~\cite{Garnett:etal:ICCV2019} by a large margin, from 5\% to 18\% in F-score and AP.

%in abstract domains
\subsection{Whole system evaluation}

We conclude our experiments with the whole system comparison between our two-stage Gen-LaneNet with prior state-of-the-art 3D-LaneNet~\cite{Garnett:etal:ICCV2019}. The apple-to-apple comparisons have been taken on all the three splits of dataset, as shown in Table~\ref{tab:whole}. On the {\it balanced scenes} the 3D-LaneNet works well, but our Gen-LaneNet  still achieves  0.8\% AP and 1.7\% F-score improvement. Considering this data split is well balanced between training and testing data and covers various scenes, it means the proposed Gen-LaneNet have better generalization on various scenes; On the  {\it rarely observed scenes}, both AP and F-score are improved 6\% and 4.4\% respectively by our method, demonstrating the superior robustness of our method when it meets uncommon test scenes; Finally on the {\it scenes with visual variations}, our method significantly surpasses the 3D-LaneNet by around 13\% in F-score and AP, which shows that our two-stage algorithm successfully benefits from the decoupled learning of the image encoding and 3D geometry reasoning. For any specific scene, we could annotate more cost-effective 2D lanes in image, to learn a general segmentation subnetwork while label a limited number of expensive 3D lanes to learn the 3D lane geometry. This makes our method a more scalable solution in real-world application. Additional qualitative comparisons are presented in Appendix~\ref{sec:visual:results}.

Besides F-score and AP, errors (euclidean distance) in meters over those matched lanes are respectively reported for {\bf near range}(0-40m) and {\bf far range}(40-100m). As observed, Gen-LaneNet maintains the error lower or on par with 3D-LaneNet, even more matched lanes are involved\footnote{A method with higher F-score considers more matched pairs of lanes to calculate the euclidean distance.}.

\section{Conclusion}
\label{sec:discuss}
We present a generalized and scalable 3D lane detection method, Gen-LaneNet. A geometry-guided anchor representation has been introduced together with a two-stage framework decoupling the learning of image segmentation and 3D lane prediction. Moreover, we present a new strategy to construct synthetic dataset for 3D lane detection. We experimentally substantiate that our method surpasses 3D-LaneNet significantly in both AP and in F-score from various perspectives.

%We show how the understanding of geometry can help design network architectures that generalize better to rarely observed scenes and texture variation. But meanwhile, there are still a lot unsolved problems aining. For example, how to solve the problem due to the limited view range associated with top-view, and how to design more appropriate network architecture to encode the geometry are both worth further study in the further.  
 
 %replaceeverything below with strong words: 0. Saying that we havetwo contributions.  1.  Contribution indataset:  uniqueness?2.Contribution part 2: Model: better in term s of precision?calculation time?  resources used?  extenability?  
\newpage
{\small
\bibliographystyle{ieee_fullname}
\bibliography{egbib}
}

%%%%%%%%%%%%%%%%%%%%%%%%%%%%%%%%%%%%%%%
%%%%%%%%%%%%%%%%%%%%%%%%%%%%%%%%%%%%%%%
%%%%%%%%%%%%%%%%%%%%%%%%%%%%%%%%%%%%%%%
%%%%%%%%%%%%%%%%%%%%%%%%%%%%%%%%%%%%%%%
%%%%%%%%%%%%%%%%%%%%%%%%%%%%%%%%%%%%%%%

\cleardoublepage

\appendix

%%%%%%%%%%%%%%%%%%%%%%%%%%%%%%%%%%%%%%%

\section{Features investigation of 3D-LaneNet}
\label{sec:vis:feat}

\begin{figure}[!h]
  \centering
  \includegraphics[width=0.48\textwidth]{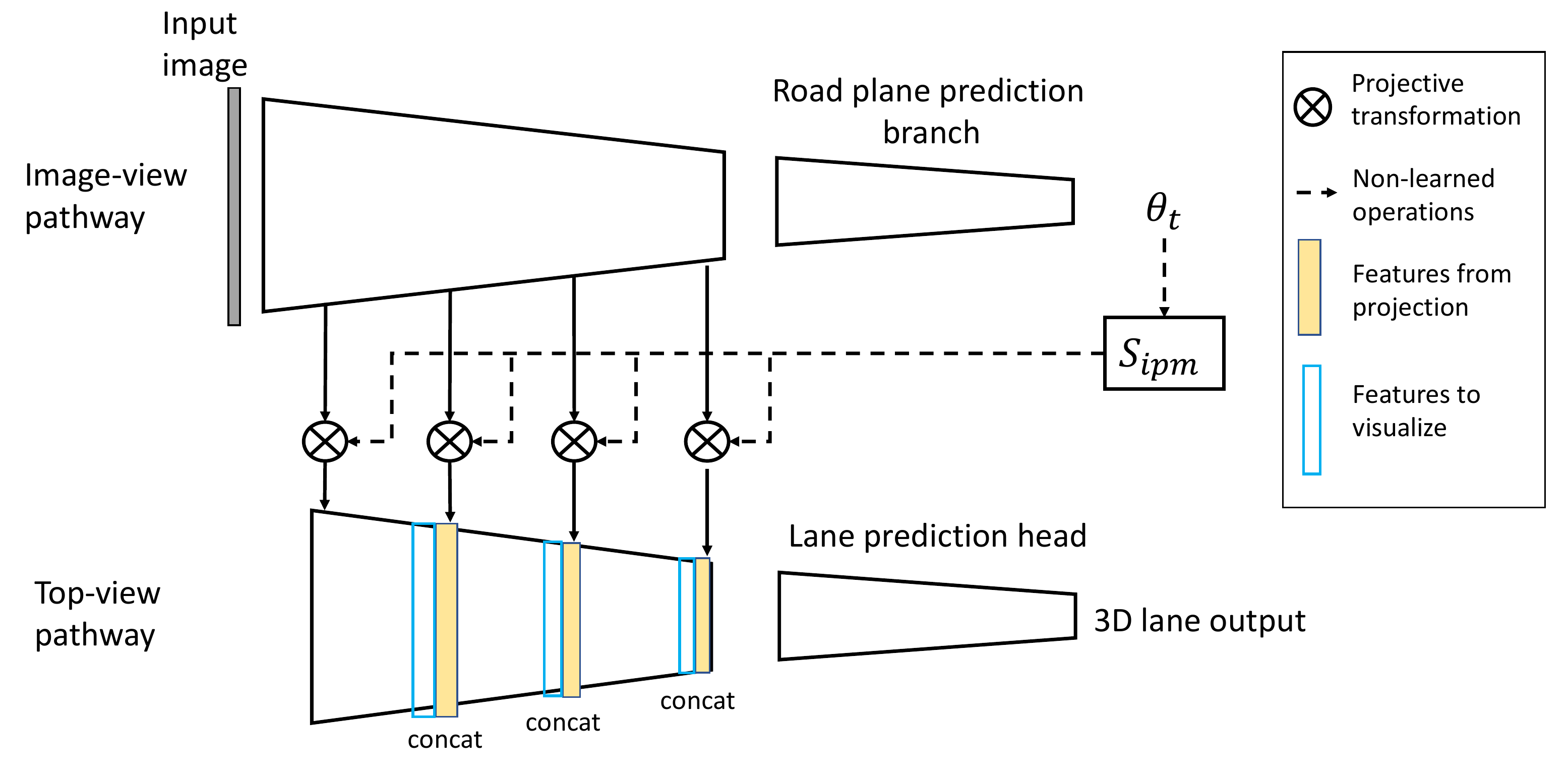}
  %\includegraphics[width=0.24\textwidth]{figs/3DLaneNet_Anchor.png} \\
  %\hspace{100pt}(a) \hspace{230pt} (b)\\
\caption{\textbf{3D-LaneNet:} An overview pipeline is shown. The whole network can be decomposed into four sub-networks: image-view pathway, road plane predication branch, top-view pathway and lane prediction head.} % (b) A figure borrowed from the 3D-LaneNet~\cite{Garnett:etal:ICCV2019} is used to illustrate the anchor-based lane representation. The anchors are defined as $N$ equally spaced vertical lines in x-positions $\{X^i_A \}^N_{i=1}$. Given a set of pre-defined fixed y-positions $\{y_i\}^K_{j=1}$, a 3D lane can be represented with an anchor $X^i_A$ composed of $2 \cdot K$ attributes $\{(x^i_j, z^i_j)\}^K_{j=1}$. A ground-truth lane is associated with its closest anchor based on x-value at $Y_{ref}$.}
  \label{fig:3DLaneNet:Arch}
\end{figure}

As we have mentioned in the paper, 3D-LaneNet~\cite{Garnett:etal:ICCV2019} represent anchor points in an inappropriate coordinate frame such that visual features can not be aligned with the prepared ground-truth lane line. We further verify this problem by investigating the key features.

As illustrated in Figure~\ref{fig:3DLaneNet:Arch}, the network of 3D-LaneNet processes information in two pathways: The \textit{image-view pathway} processes and preserves information from the image while the \textit{top-view pathway} processes features in top-view and uses them to predict the 3D lane output. Information flows from image-view pathway to the top-view pathway through four projective transformation layers. 

\begin{figure}[!h]
  \centering
  \includegraphics[height=0.14\textwidth]{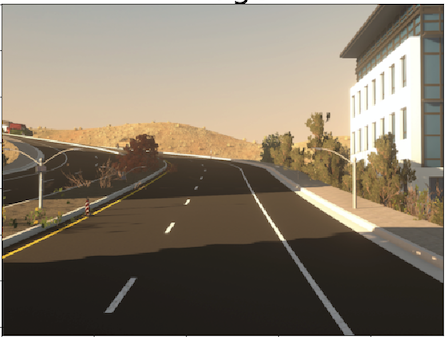}
  \includegraphics[height=0.14\textwidth]{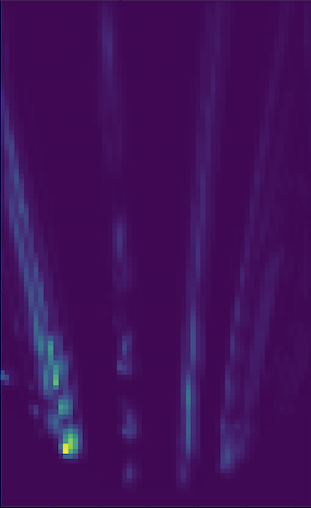}
  \includegraphics[height=0.14\textwidth]{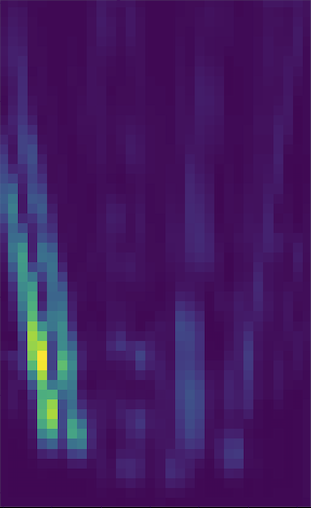}
  \includegraphics[height=0.14\textwidth]{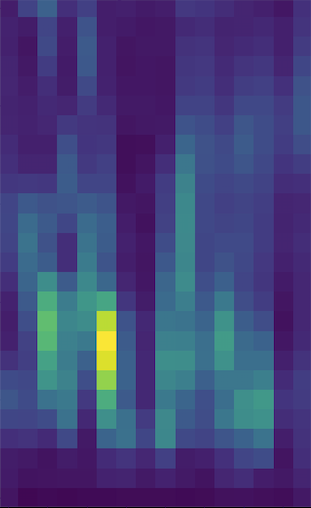}
\caption{\textbf{3D-LaneNet feature visualization:} Given an image captured on a uphill road, imaged lanes are suppose to appear diverging when projected to the virtual top-view. As observed from the visualized features from those blue-marked layers in Figure~\ref{fig:3DLaneNet:Arch}(a), \textbf{top-view features} also form diverging lines. Consequently, diverging visual features will not be in alignment with the parallel lane lines prepared as ground-truth.}
  \label{fig:feat:vis}
\end{figure}

To confirm the alignment between top-view features and the ground-truth lane, we choose to visualize feature from a few key layers of the top-view pathway, which are marked in blue in Figure~\ref{fig:3DLaneNet:Arch}. As illustrated in Figure~\ref{fig:feat:vis}, for a uphill road, image lanes projected to the virtual top-view are expected to appear diverging rather than parallel with each other. When the features from the key layers are visualized, we can observe the same diverging appearance in the feature space. However, the anchor representation from 3D-LaneNet would provide parallel ground-truth lines, which could not align with the diverging features. Although the network learns to focus on lanes, where features are high-lighted, the network can not deform the features to their targeting positions internally. As a result, the misalignment between visual features and ground truth makes the method not generalizable to unobserved scenes.

\section{Algebraic derivation of the geometric transformation}
\label{sec:geo:derive}

In this section, we present an algebraic derivation of the geometric transformation between 3D ego-vehicle coordinate frame and virtual top-view coordinate frame. Although a general derivation from geometric perspective has been presented in our main paper, we present this algebraic derivation as a double-verification. The derivation considers a simpler camera setup when only pitch angle is involved in camera orientation.

A 3D point $(x, y, z)$ in ego-vehicle coordinate frame can be projected to a 2D image point $(u, v)$ through a projective transformation. Its corresponding 2D point $(\bar{x}, \bar{y})$ in top-view coordinate frame can be projected to the same 2D image point through a planer homography. Given $R$ as the rotation matrix, $T$ as the translation vector, $K$ indicating camera intrinsic parameters, the described relationship can be written as: 
{\small
\begin{gather}
\label{eqn:geo}
    \begin{bmatrix} u \\ v \\ 1 \end{bmatrix}
     =
    \alpha_1 K
    \begin{bmatrix}
        R & T
    \end{bmatrix}
    \begin{bmatrix}
        x \\ y \\ z \\ 1
    \end{bmatrix}
    =
    \alpha_2 K
    \begin{bmatrix}
        R_{1:2} & T
    \end{bmatrix}
    \begin{bmatrix}
        \bar{x} \\ \bar{y} \\ 1
    \end{bmatrix},
\end{gather}
}
\noindent where $R_{1:2}$ indicates the first two columns of $R$, and $\alpha_1$, $\alpha_2$ are two constant coefficients. Given camera height $h$, pitch angle $\theta$, we can explicitly write $R, T$ as: 
{\small
\begin{gather}
    R
     =
    \begin{bmatrix}
        1 &             0 &             0\\
        0 & -\sin{\theta} & -\cos{\theta}\\
        0 &  \cos{\theta} & -\sin{\theta}\\
    \end{bmatrix},
    T =
    \begin{bmatrix}
        0 \\ \cos{\theta} \cdot h \\ \sin{\theta} \cdot h \\
    \end{bmatrix}.
\end{gather}
}
\noindent Given simplified notation of $\sin{\theta}$ as $s$,  $\cos{\theta}$ as $c$, we can rewrite Equation~\ref{eqn:geo} as:
{\small
\begin{gather}
    \begin{bmatrix}
        1 &  0 &  0 & 0\\
        0 & -s & -c & ch\\
        0 &  c & -s & sh\\
    \end{bmatrix}
    \begin{bmatrix}
         x \\ y \\ z \\ 1
    \end{bmatrix}
    =
    \alpha 
    \begin{bmatrix}
        1 &  0 &  0\\
        0 & -s & ch\\
        0 &  c & sh\\
    \end{bmatrix}
    \begin{bmatrix}
       \bar{x} \\ \bar{y} \\ 1
    \end{bmatrix}.  \nonumber
\end{gather}
}
\noindent This equitation can expended as:
{\small
\begin{gather}
    x
    \begin{bmatrix}
    1 \\ 0 \\ 0
    \end{bmatrix} + y
    \begin{bmatrix}
    0 \\ -s \\ c
    \end{bmatrix} + z
    \begin{bmatrix}
    0 \\ -c \\ -s
    \end{bmatrix} + 
    \begin{bmatrix}
    0 \\ ch \\ sh
    \end{bmatrix} = \nonumber\\
    \alpha \bar{x}
    \begin{bmatrix}
    1 \\ 0 \\ 0
    \end{bmatrix} + \alpha \bar{y}
    \begin{bmatrix}
    0 \\ -s \\ c
    \end{bmatrix} + \alpha
    \begin{bmatrix}
    0 \\ ch \\ sh
    \end{bmatrix}, \nonumber
\end{gather}
}
\noindent which further leads to three equations in scalars:
\begin{eqnarray}
    x &=& \alpha \bar{x} \label{eqn:1} \\
    -sy - cz + ch &=& -\alpha s \bar{y} + \alpha ch \label{eqn:2}\\
    cy - sz + sh &=& \alpha c \bar{y} + \alpha sh \label{eqn:3}.
\end{eqnarray}
Reorganizing Equation~\ref{eqn:3} in
\begin{equation}
    y = \alpha \bar{y} + \frac{s}{c}(\alpha h - h + z), \label{eqn:4}
\end{equation}
\noindent and substituting Equation~\ref{eqn:4} with $y$ in Equation~\ref{eqn:2}, we derive $\alpha$ step-by-step:
\\
\begin{eqnarray}
    -\alpha s \bar{y} + \alpha c h = -s \big[ \alpha \bar{y} + \frac{s}{c}(\alpha h - h + z) \big] - c z + ch \nonumber \\
   {\color{blue}\Longrightarrow} \alpha c h = - \alpha \frac{s^2 h}{c} + \frac{s^2 (h-z)}{c} + c(h-z) \nonumber\\
   {\color{blue}\Longrightarrow }\space\space   \frac{s^2 + c^2}{c} \alpha h = \frac{s^2 + c^2}{c} (h-z) \nonumber\\
   {\color{blue}\Longrightarrow \space\space } \alpha = \frac{h-z}{h} \label{eqn:5}
\end{eqnarray}
\\
Substituting Equation~\ref{eqn:5} with $\alpha$ in Equation~\ref{eqn:1} and Equation~\ref{eqn:4} respectively, we at last derive the equations:
\begin{eqnarray}
    \label{eqn:sim_geo1}
        x &=& \bar{x} \cdot \frac{h - z}{h}\\ 
    \label{eqn:sim_geo2}
        y &=& \bar{y} \cdot \frac{h - z}{h},
\end{eqnarray}
So far we have algebraically derived the geometric transformation between 3D coordinate frame and virtual top-view coordinate frame. The transformation agrees with the geometric proof presented in the main paper.

%%%%%%%%%%%%%%%%%%%%%%%%%%%%%%%%%%%%%%%%%%%%%%%%%%%%%%%%%%%%%%%%%%%%%%%%%%%%%%%%%%%%%%%5

\section{Experiments on center lines}
\label{sec:exp:centerline}

%We evaluate on the simulation dataset presented in our paper. We have not tested on either the simulated dataset or the real dataset presented in~\cite{Garnett:etal:ICCV2019}, because at the time we were submitting this paper, neither of these datasets was available. It is worth mentioning that we have not included lane merge/split examples in our experiments due to lack of such rare samples. We agree the anchor design presented in 3D-LaneNet~\cite{Garnett:etal:ICCV2019} considering multiple center lanes in a single anchor can handle merge/split cases reasonably well.

Similar to the evaluation of lane line prediction, we conduct evaluation on center line prediction from three perspectives: the effect of new anchor representation, Table~\ref{tab:anchor:compare:centerline}; the upper bound of two-stage framework, Table~\ref{tab:twostage:centerline}; and the whole system comparison, Table~\ref{tab:whole:centerline}. A candidate method is also evaluated under three different splits of dataset: {\it Balanced scenes}, {\it Rarely observed scenes}, and {\it Scenes with visual variations}. Observe from the evaluation of center line prediction, similar conclusion can be drawn compared to the evaluation of the lane line prediction.

%%%%%%%%%%%%%%%%%%%%%%%%%%%%%%%%%%%%%%%%%%%%%%%%%%%%%%%%%%%%%%%%%%%%%%%%%%%%%%%%%%%%%%%5

\begin{table*}[]
\centering
\begin{tabular}{|c|c|c|c|c|c|c|c|c|c|c|}
\hline
\multicolumn{2}{|c|}{}                                                                                   & \multicolumn{3}{c|}{\textbf{balanced  scenes}} & \multicolumn{3}{c|}{\textbf{rarely observed}} & \multicolumn{3}{c|}{\textbf{visual variations}} \\ \cline{3-11} 
\multicolumn{2}{|c|}{\multirow{-2}{*}{\textbf{\begin{tabular}[c]{@{}c@{}}\end{tabular}}}} & w/o                  & w/                   & gain                                       & w/o                    & w/                     & gain                                          & w/o                      & w/                       & gain                                             \\ \hline
                                                               & \textbf{F-score}                        & 89.5                 & 93.3                 & {\color[HTML]{FE0000} +3.8}                & 77.0                   & 84.1                   & {\color[HTML]{FE0000} +7.1}                   & 75.5                     & 86.6                     & {\color[HTML]{FE0000} +11.1}                     \\ \cline{2-11} 
\multirow{-2}{*}{\textbf{3D-LaneNet}}                          & \textbf{AP}                             & 91.4                 & 95.5                 & {\color[HTML]{FE0000} +4.1}                & 80.0                   & 85.9                   & {\color[HTML]{FE0000} +5.9}                   & 77.7                     & 88.7                     & {\color[HTML]{FE0000} +11.0}                      \\ \hline
                                                               & \textbf{F-score}                        & 91.2                 & 94.5                 & {\color[HTML]{FE0000} +3.3}                & 79.7                   & 85.9                   & {\color[HTML]{FE0000} +6.2}                   & 87.9                     & 92.3                     & {\color[HTML]{FE0000} +4.4}                      \\ \cline{2-11} 
\multirow{-2}{*}{\textbf{3D-GeoNet}}                           & \textbf{AP}                             & 93.2                 & 96.8                 & {\color[HTML]{FE0000} +3.6}                & 83.0                   & 87.7                   & {\color[HTML]{FE0000} +4.7}                   & 90.6                     & 94.2                     & {\color[HTML]{FE0000} +3.6}                      \\ \hline
                                                               & \textbf{F-score}                        & 88.2                 & 90.8                 & {\color[HTML]{FE0000} +2.6}                & 76.1                   & 79.5                   & {\color[HTML]{FE0000} +3.4}                   & 84.2                     & 88.2                     & {\color[HTML]{FE0000} +4.0}                      \\ \cline{2-11} 
\multirow{-2}{*}{\textbf{Gen-LaneNet}}                         & \textbf{AP}                             & 90.8                 & 92.6                 & {\color[HTML]{FE0000} +1.8}                & 79.4                   & 80.6                   & {\color[HTML]{FE0000} +1.2}                   & 87.0                     & 90.0                     & {\color[HTML]{FE0000} +3.0}                      \\ \hline
\end{tabular}
\caption{(Center line) Comparison of anchor representations. "w/o" represents the integration with anchor design in~\cite{Garnett:etal:ICCV2019}, while "w" represents the integration with our anchor design. For convenience, we also shows the performance gain by integrating our anchor design.}
\label{tab:anchor:compare:centerline}
\end{table*}

\textbf{Anchor effect:} As shown in Table~\ref{tab:anchor:compare:centerline}, the introduction of our new anchor leads to consistent improvement over all candidate methods and over all splits of dataset. Substantial improvement can be observed on 3D-LaneNet on {\it rarely observed scenes} and {\it scenes with visual variations} with 7.1\% and 11.1\% improvements in F-score respectively. This observation verifies the importance of the new anchor and prove that establishing alignment between visual features and lane labels help generalization to unobserved scenes of visual appearance.

%%%%%%%%%%%%%%%%%%%%%%%%%%%%%%%%%%%%%%%%%%%%%%%%%%%%%%%%%%%%%%%%%%%%%%%%%%%%%%%%%%%%%%%5

\begin{table*}[]
\centering
\begin{tabular}{|c|c|c|c|c|}
\hline
\multicolumn{2}{|c|}{\textbf{}} & \textbf{balanced  scenes} & \textbf{rarely observed} & \textbf{visual variations} \\ \hline
 & \textbf{F-score} & 89.5 & 77.0 & 75.5 \\ \cline{2-5} 
\multirow{-2}{*}{\textbf{3D-LaneNet}} & \textbf{AP} & 91.4 & 80.0 & 77.7 \\ \hline
 & \textbf{F-score} & {\color[HTML]{3166FF} \textbf{94.5}} & {\color[HTML]{3166FF} \textbf{85.9}} & {\color[HTML]{3166FF} \textbf{92.3}} \\ \cline{2-5} 
\multirow{-2}{*}{\textbf{3D-GeoNet}} & \textbf{AP} & {\color[HTML]{3166FF} \textbf{96.8}} & {\color[HTML]{3166FF} \textbf{87.7}} & {\color[HTML]{3166FF} \textbf{94.2}} \\ \hline
 & \textbf{F-score} & \textbf{90.8} & \textbf{79.5} & \textbf{88.2} \\ \cline{2-5} 
\multirow{-2}{*}{\textbf{Gen-LaneNet}} & \textbf{AP} & \textbf{92.6} & \textbf{80.6} & \textbf{90.0} \\ \hline
\end{tabular}
\caption{(Center line) Upper bound of the proposed two-stage framework. 3D-GeoNet shows potential improvement on Gen-LaneNet when a better image segmentation algorithm is integrated.}
\label{tab:twostage:centerline}
\end{table*}

\textbf{The upper bound of two-stage framework:} As shown in Table~\ref{tab:twostage:centerline}, the two-stage framework Gen-LaneNet appears superior to the end-to-end learned method 3D-LaneNet in all three splits of dataset. 3D-GeoNet achieve the highest performance in all cases which shed a light on the upper bound of Gen-LaneNet given perfect image segmentation. Specifically, the margin between 3D-LaneNet and 3D-GeoNet can be significantly large, 16\% in both F-score and AP, on scenes with visual variations. Meanwhile, Gen-LaneNet is shown to gain significantly when provided with more available 2D labels and better segmentation network.

\textbf{Whole system comparison:} As observed from Table~\ref{tab:whole:centerline}, Gen-LaneNet surpasses 3D-LaneNet over all splits of dataset. The most significant improvement appears under scenes with visual variations (13\% in both AP and F-score), where the 3D labels have not included certain illumination but 2D labels have. Besides F-score and AP, x, z errors from close range (0-40 m) and far range (40 - 80 m) are also reported. Although Gen-LaneNet compute these errors over more matched pairs of predicted lanes and ground-truth lanes, the localization errors of its result are maintained lower or on par with 3D-LaneNet.

%%%%%%%%%%%%%%%%%%%%%%%%%%%%%%%%%%%%%%%%%%%%%%%%%%%%%%%%%%%%%%%%%%%%%%%%%%%%%%%%%%%%%%%5

\begin{table*}[]
\centering
\begin{tabular}{|c|c|c|c|c|c|c|c|}
\hline
\textbf{Dataset Splits} & \textbf{Method} & \textbf{F-Score} & \textbf{AP} & \textbf{\begin{tabular}[c]{@{}c@{}}x error\\ near (m)\end{tabular}} & \textbf{\begin{tabular}[c]{@{}c@{}}x error \\ far (m)\end{tabular}} & \textbf{\begin{tabular}[c]{@{}c@{}}z error\\ near (m)\end{tabular}} & \textbf{\begin{tabular}[c]{@{}c@{}}z error\\ far (m)\end{tabular}} \\ \hline
 & 3D-LaneNet & 89.5 & 91.4 & 0.066 & 0.456 & 0.015 & 0.179 \\ \cline{2-8} 
\multirow{-2}{*}{\textbf{\begin{tabular}[c]{@{}c@{}}balanced \\  scenes\end{tabular}}} & Gen-LaneNet & {\color[HTML]{FE0000} 90.8} & {\color[HTML]{FE0000} 92.6} & 0.055 & 0.457 & 0.011 & 0.176 \\ \hline
 & 3D-LaneNet & 77.0 & 80.0 & 0.162 & 0.883 & 0.040 & 0.557 \\ \cline{2-8} 
\multirow{-2}{*}{\textbf{\begin{tabular}[c]{@{}c@{}}rarely\\ observed\end{tabular}}} & Gen-LaneNet & {\color[HTML]{FE0000} 79.5} & {\color[HTML]{FE0000} 80.6} & 0.121 & 0.885 & 0.026 & 0.547 \\ \hline
 & 3D-LaneNet & 75.5 & 77.7 & 0.120 & 0.636 & 0.030 & 0.227 \\ \cline{2-8} 
\multirow{-2}{*}{\textbf{\begin{tabular}[c]{@{}c@{}} visual\\ variations\end{tabular}}} & Gen-LaneNet & {\color[HTML]{FE0000} 88.2} & {\color[HTML]{FE0000} 90.0} & 0.072 & 0.438 & 0.015 & 0.187 \\ \hline
\end{tabular}
\caption{(Center line) Whole system comparison between 3D-LaneNet~\cite{Garnett:etal:ICCV2019} and Gen-LaneNet. }
\label{tab:whole:centerline}
\end{table*}

\section{Qualitative comparison}
\label{sec:visual:results}

\begin{figure*}[!h]
  \centering
  \includegraphics[width=0.33\textwidth]{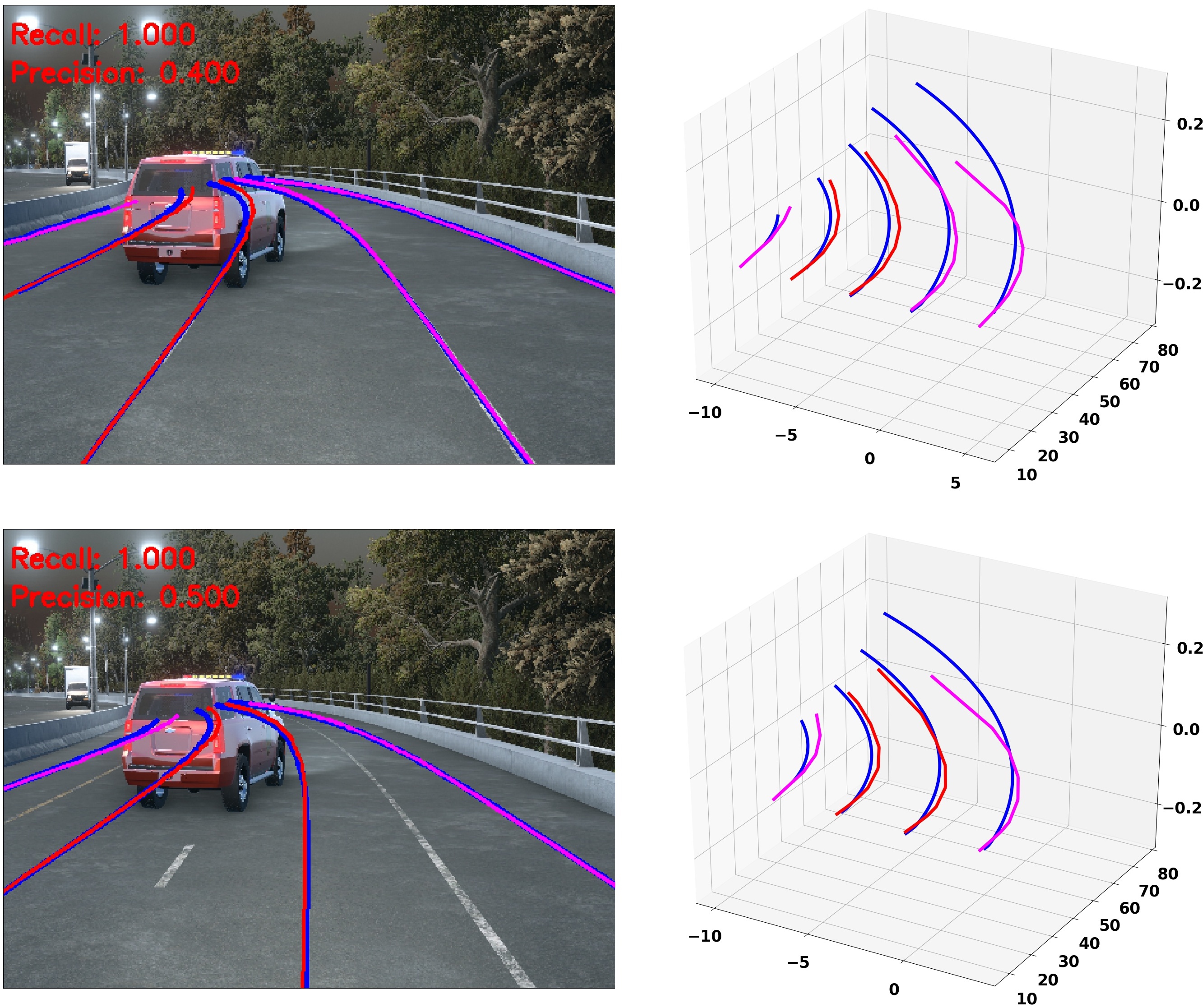} \hspace{1cm}
  \includegraphics[width=0.33\textwidth]{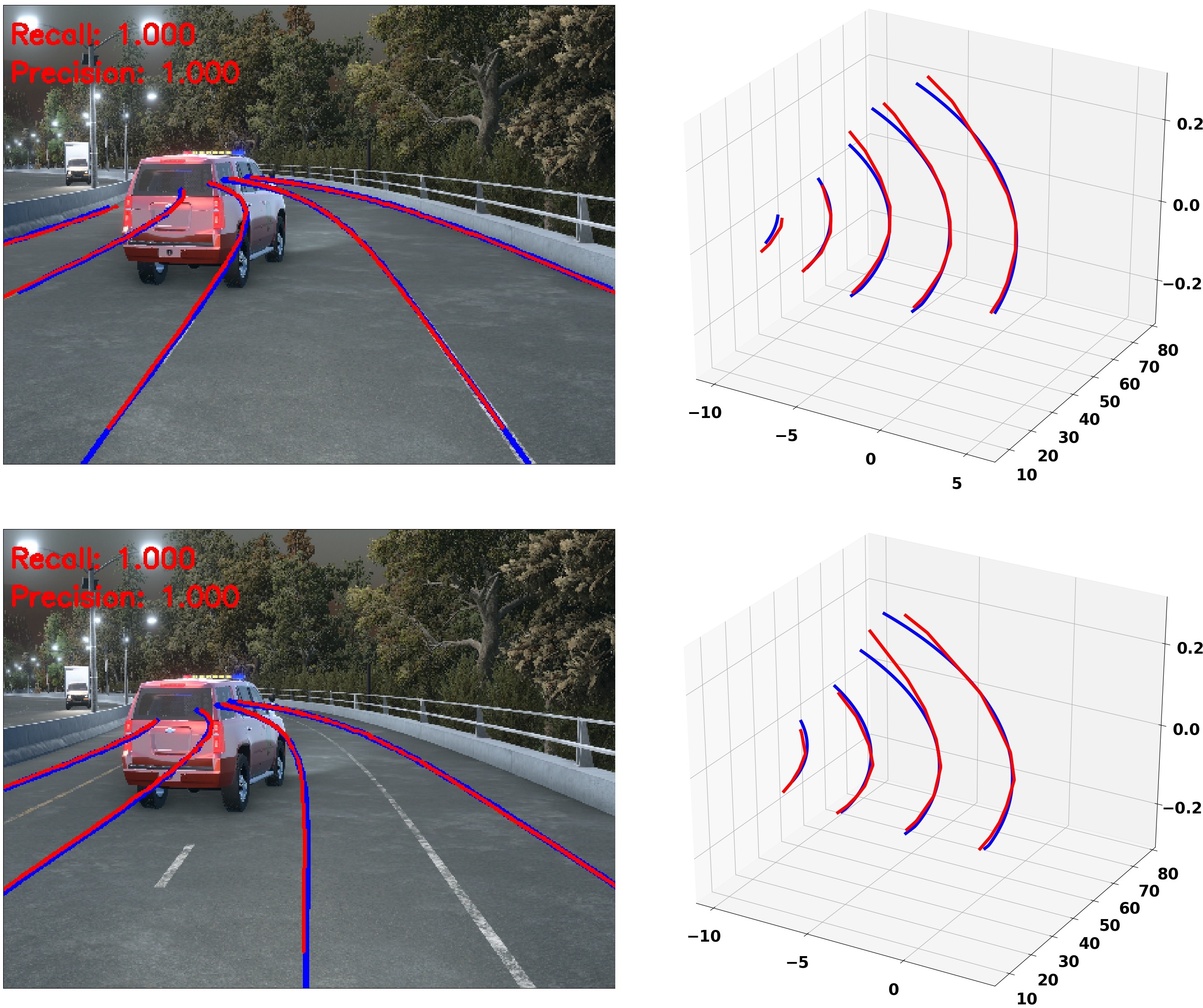}
  \HRule\\
  \includegraphics[width=0.33\textwidth]{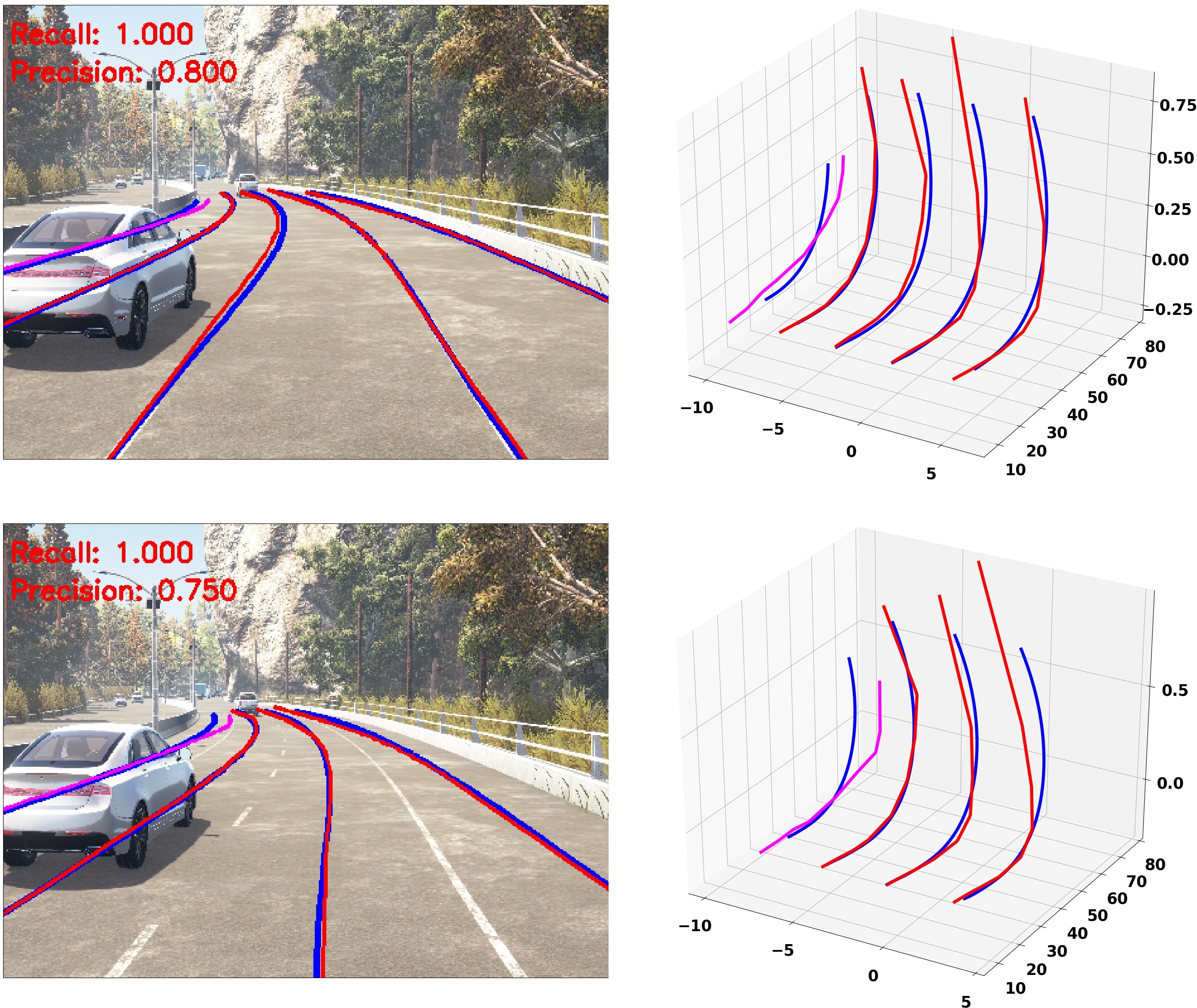} \hspace{1cm}
  \includegraphics[width=0.33\textwidth]{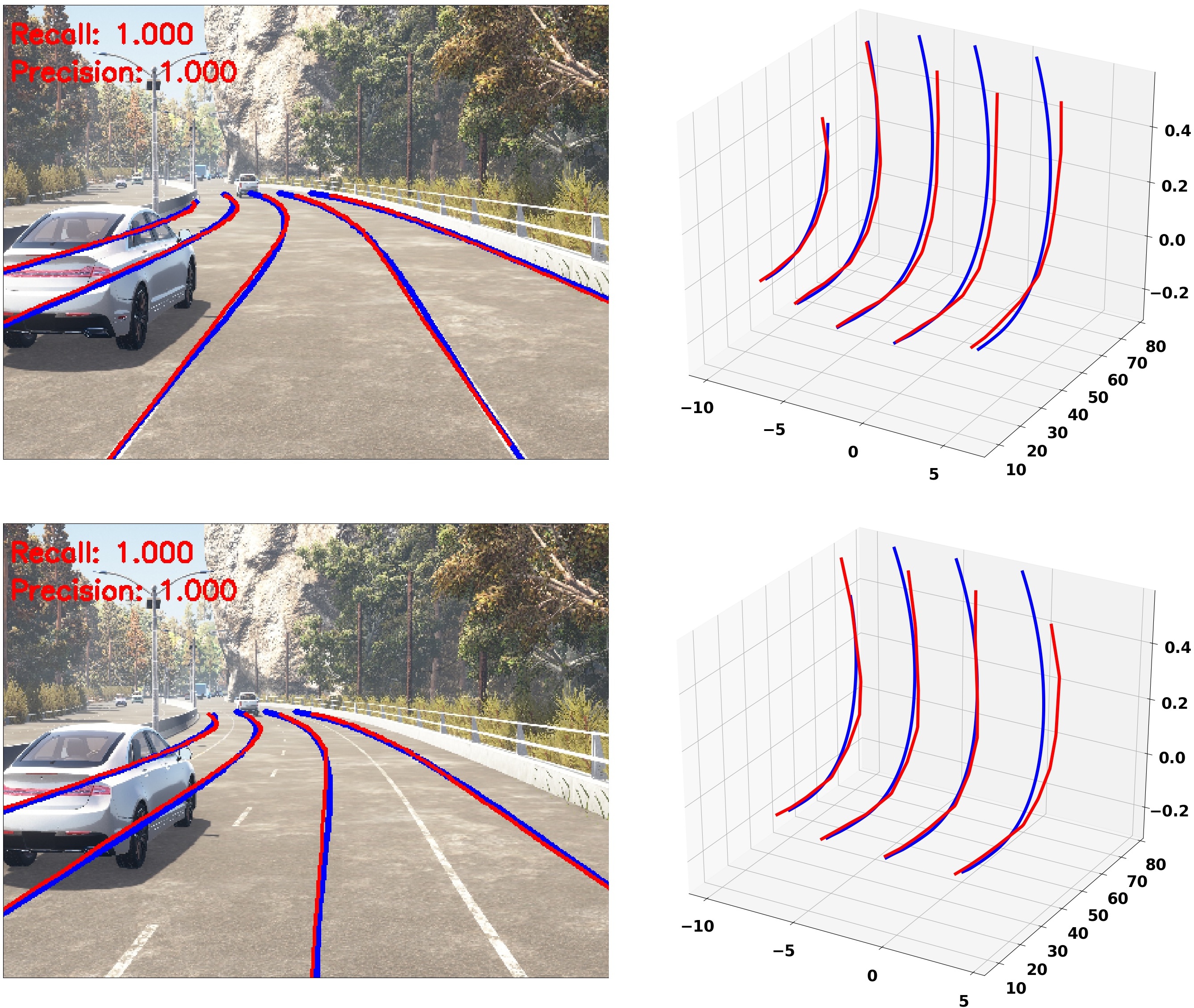}
  \HRule\\
  %\includegraphics[width=0.29\textwidth]{figs/visual_results/3DLaneNet/images_04_0000626.jpg} \hspace{1cm}
  %\includegraphics[width=0.29\textwidth]{figs/visual_results/3DLaneNetPlus/images_04_0000626.jpg}
  %\HRule\\
  \includegraphics[width=0.33\textwidth]{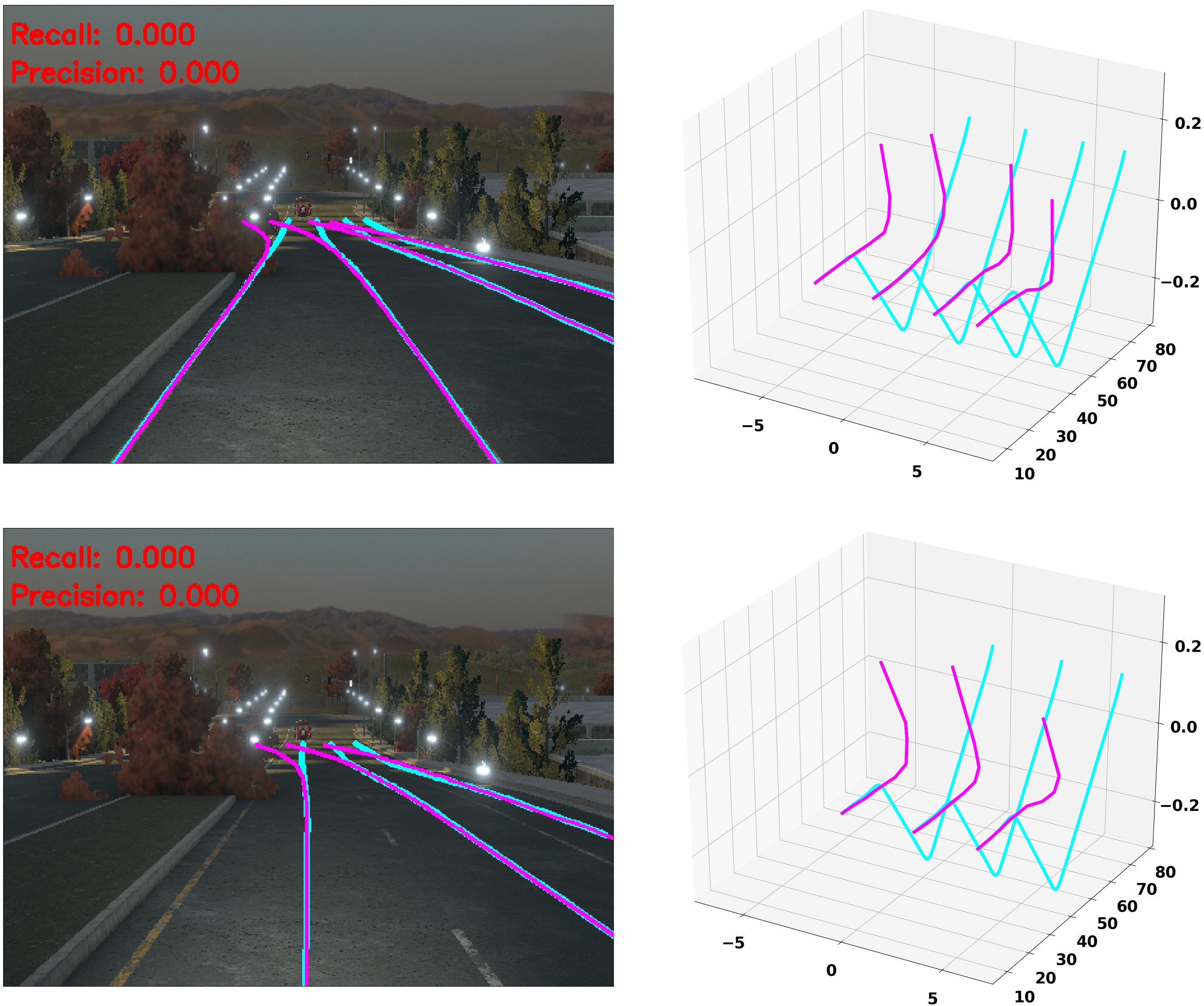} \hspace{1cm}
  \includegraphics[width=0.33\textwidth]{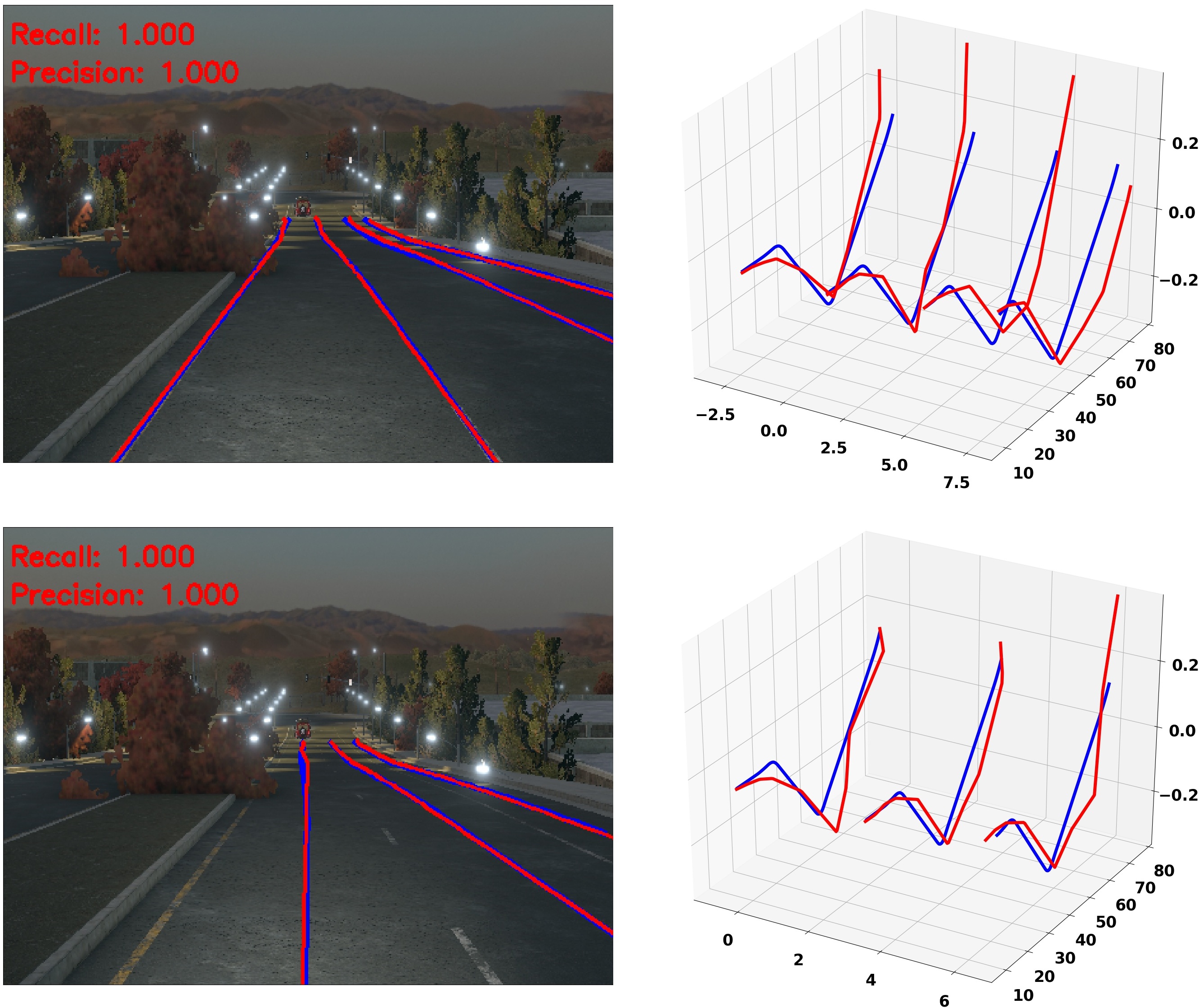}
  \HRule\\
  \includegraphics[width=0.33\textwidth]{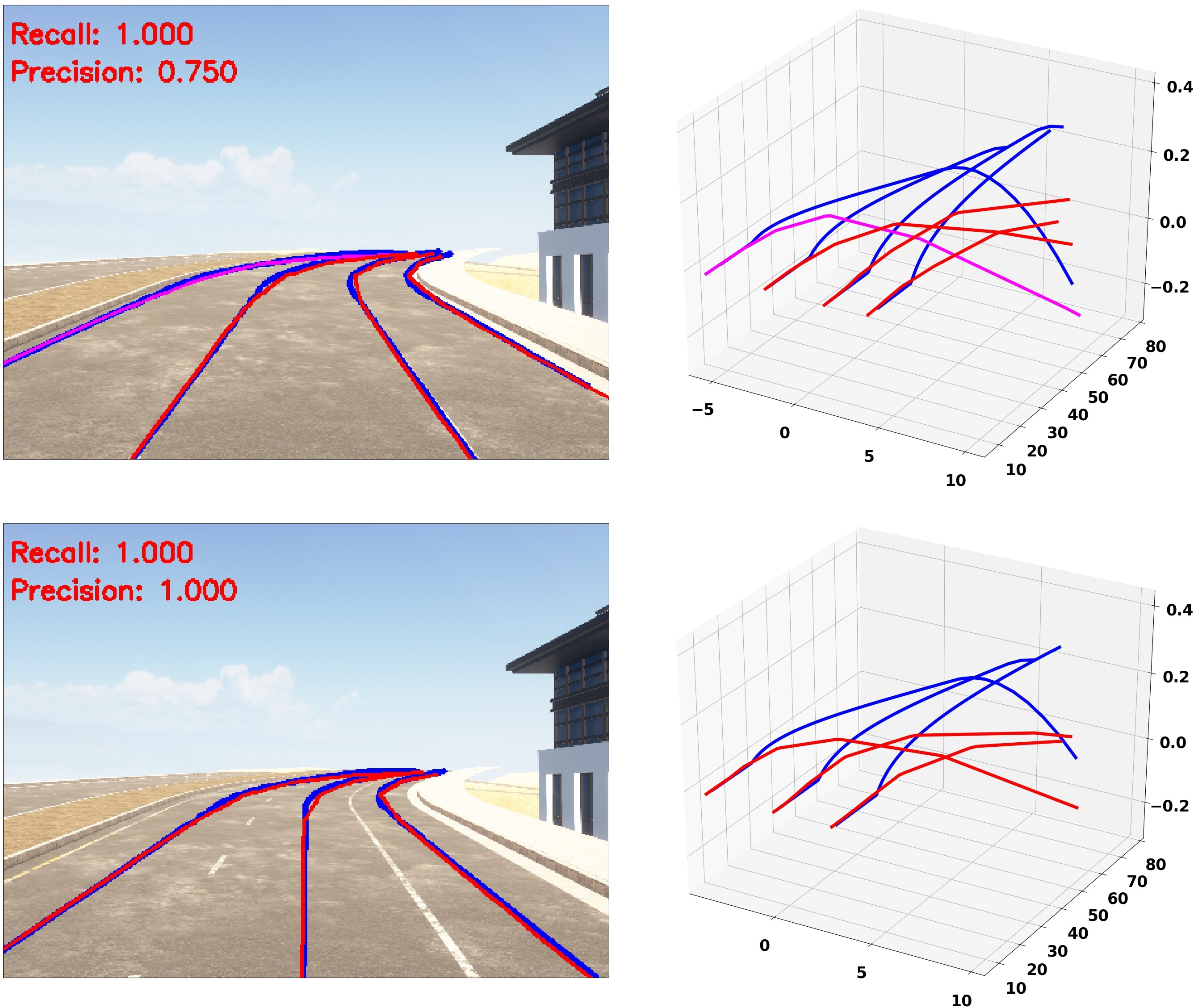} \hspace{1cm}
  \includegraphics[width=0.33\textwidth]{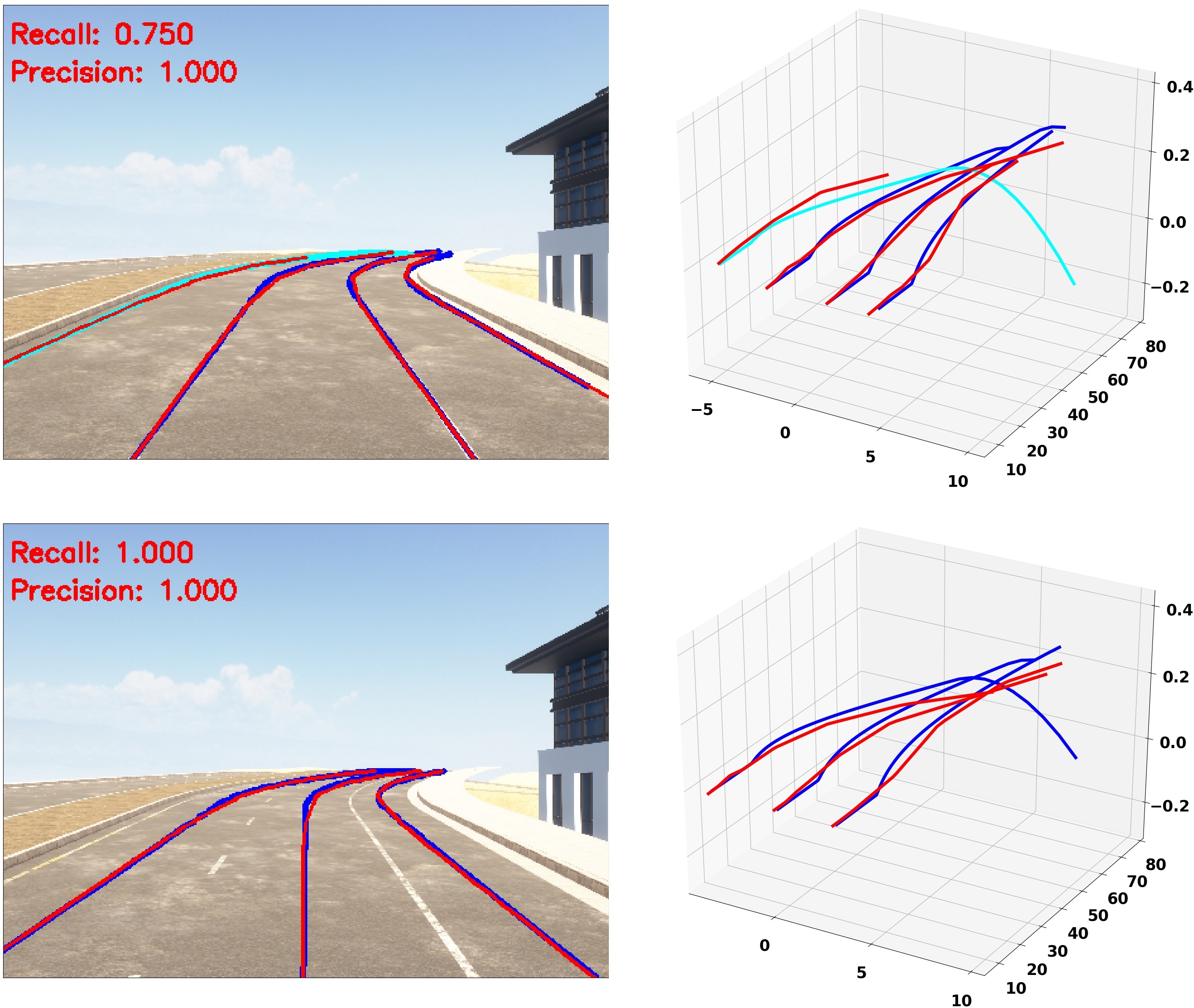}\\
    \hspace{15pt} 3D-LaneNet  \hspace{80pt} 3D-LaneNet (new anchor)
\caption{\textbf{Effect of the new anchor}. Predicted lanes from 3D-LaneNet and from the extended version with our new anchor are visually compared. Examples are chosen from the test set given the standard five-fold split of the whole dataset. Observe that adopting our new anchor consistently improves the localization of laneline and in turn leads better prediction. For each example, we show image results on the left and 3D results on the right, lane line results in the top row, and center line results in the bottom row. We color the ground-truth lanes in blue and the predicted lanes in red. While the purple lanes are missed detections and the cyan lanes are false alarms.}
  \label{fig:visual:1}
\end{figure*}

\begin{figure*}[!h]
  \centering
\includegraphics[width=0.33\textwidth]{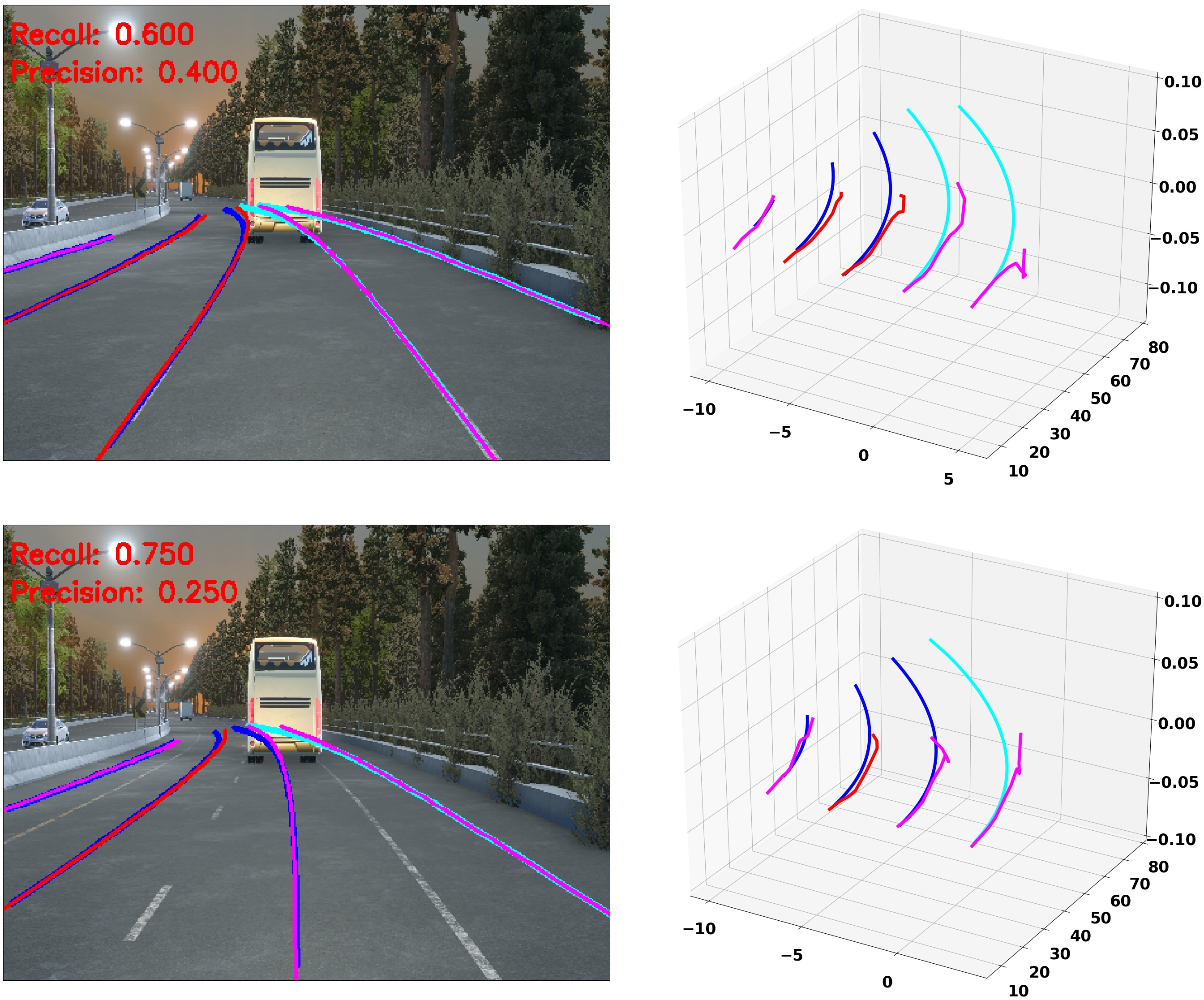} \hspace{1cm}
  \includegraphics[width=0.33\textwidth]{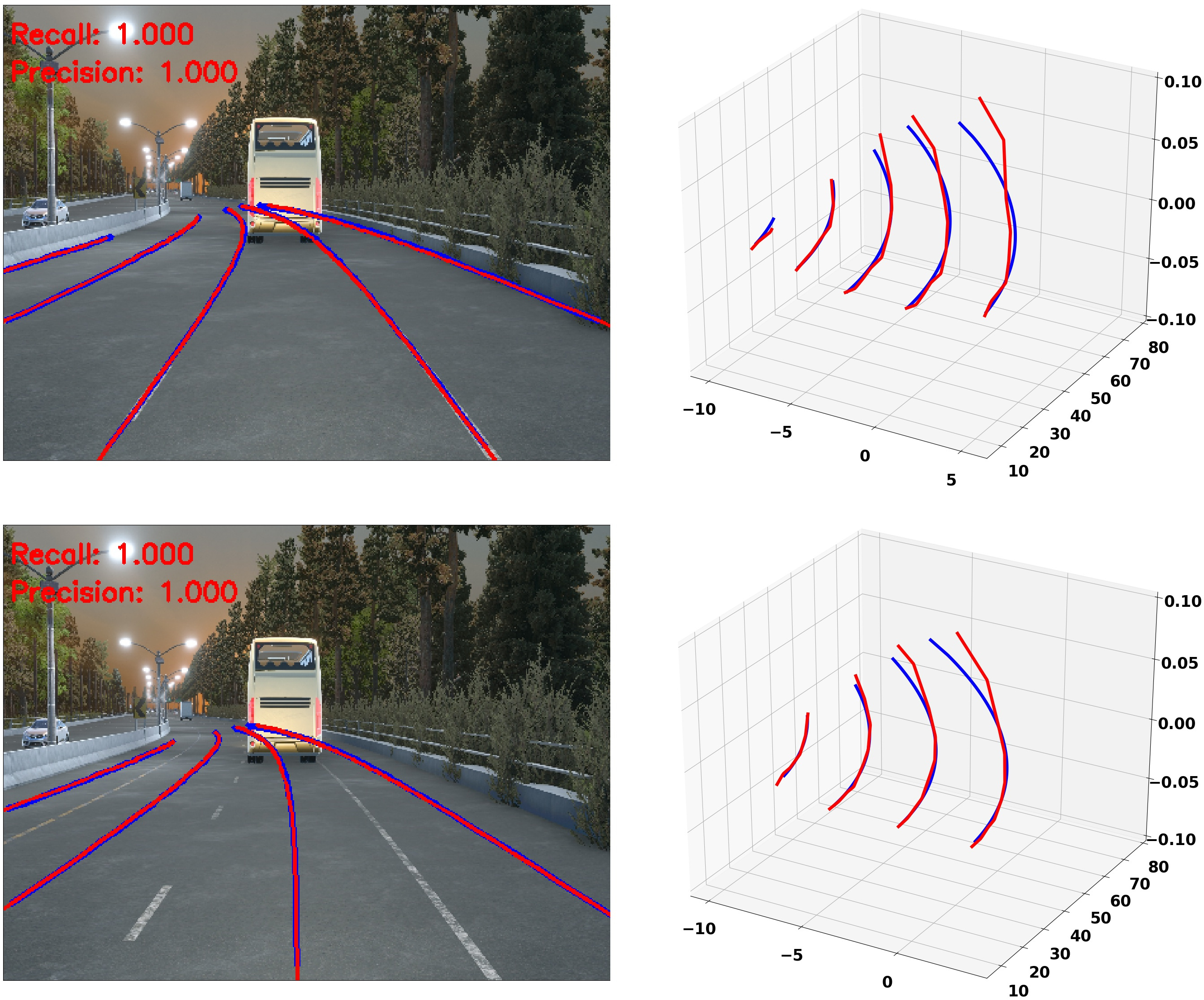}
  \HRule\\
  \includegraphics[width=0.33\textwidth]{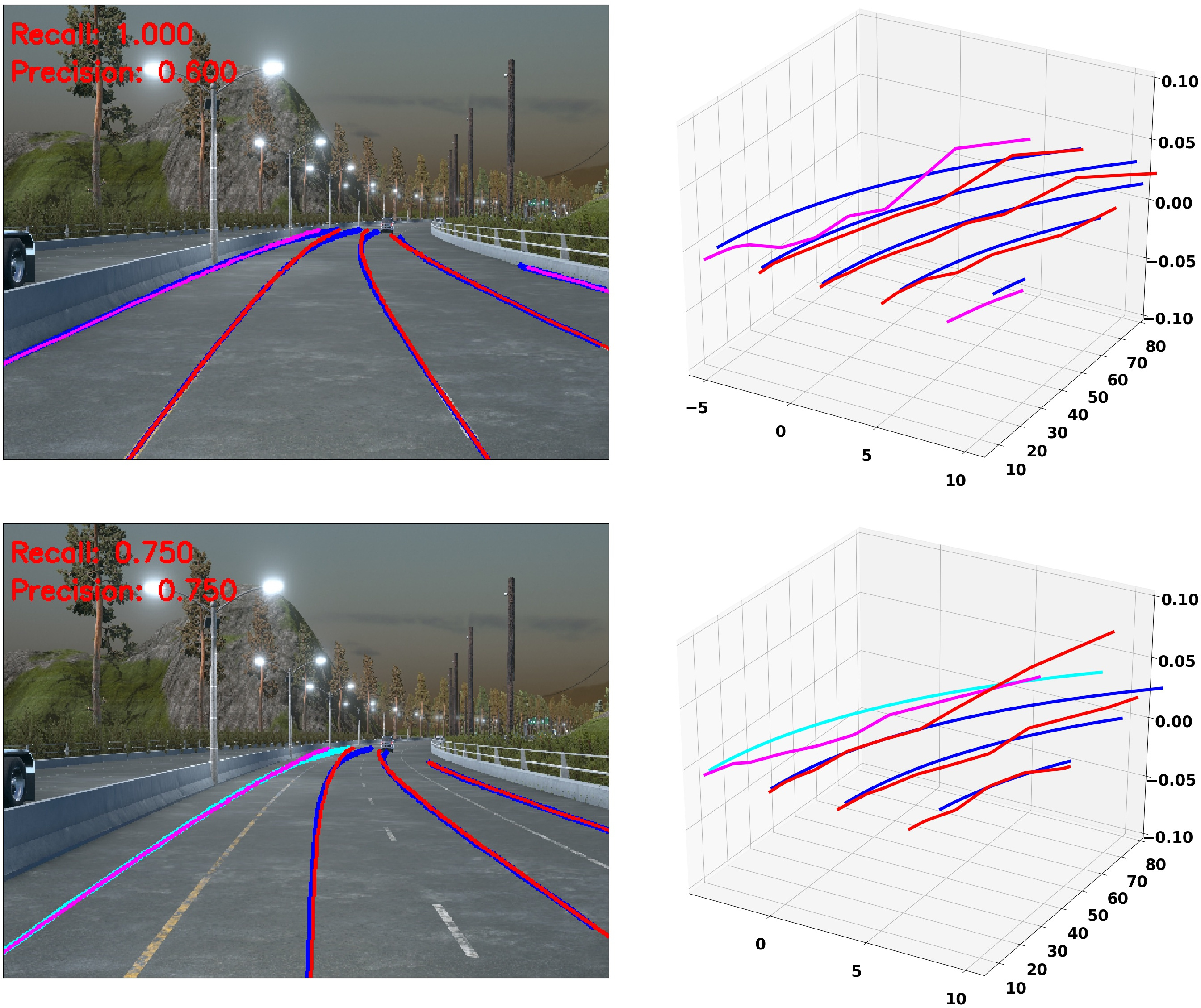} \hspace{1cm}
  \includegraphics[width=0.33\textwidth]{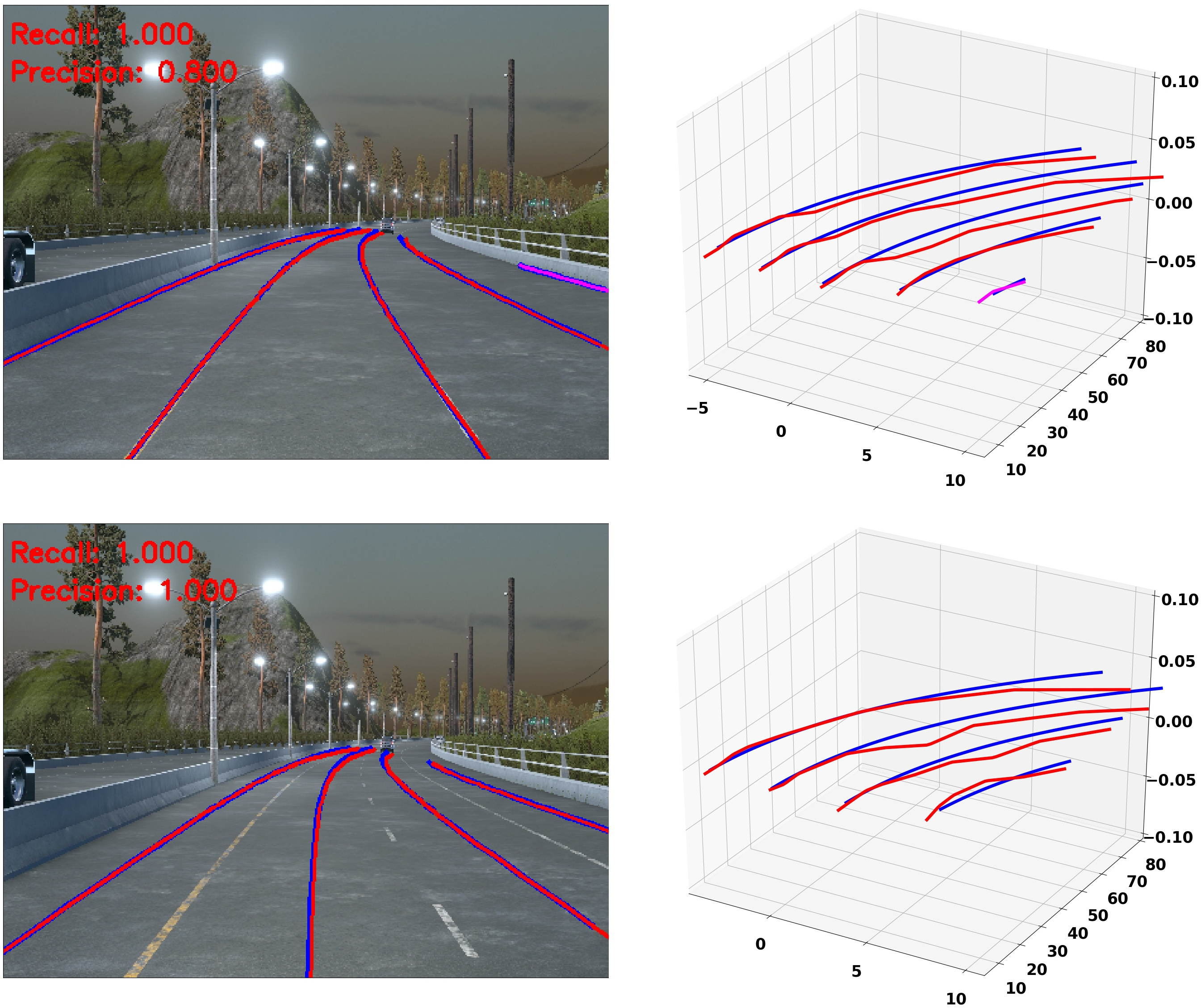}
  \HRule\\
  %\includegraphics[width=0.29\textwidth]{figs/visual_results/3DLaneNet/images_04_0000626.jpg} \hspace{1cm}
  %\includegraphics[width=0.29\textwidth]{figs/visual_results/3DLaneNetPlus/images_04_0000626.jpg}
  %\HRule\\
  \includegraphics[width=0.33\textwidth]{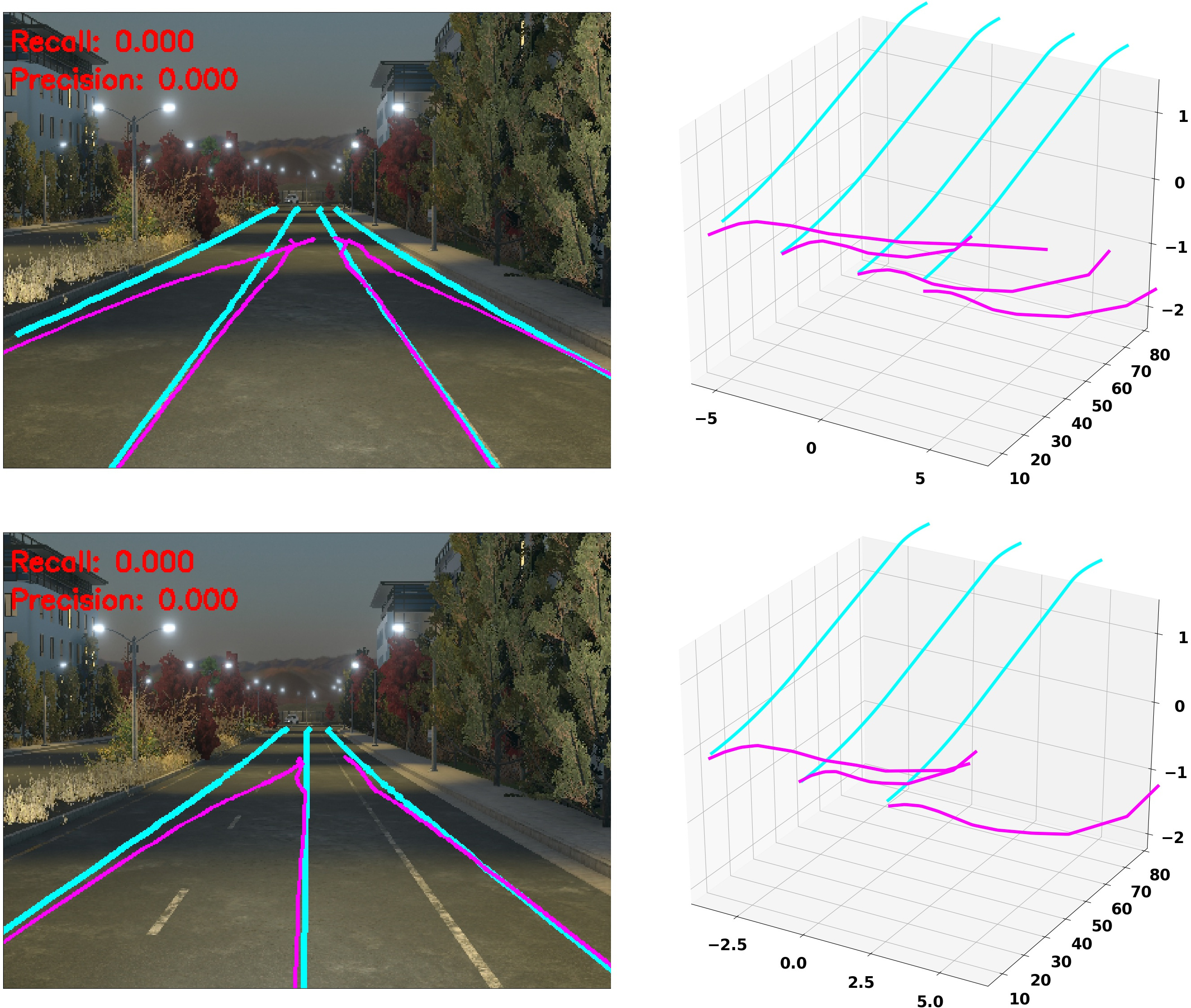} \hspace{1cm}
  \includegraphics[width=0.33\textwidth]{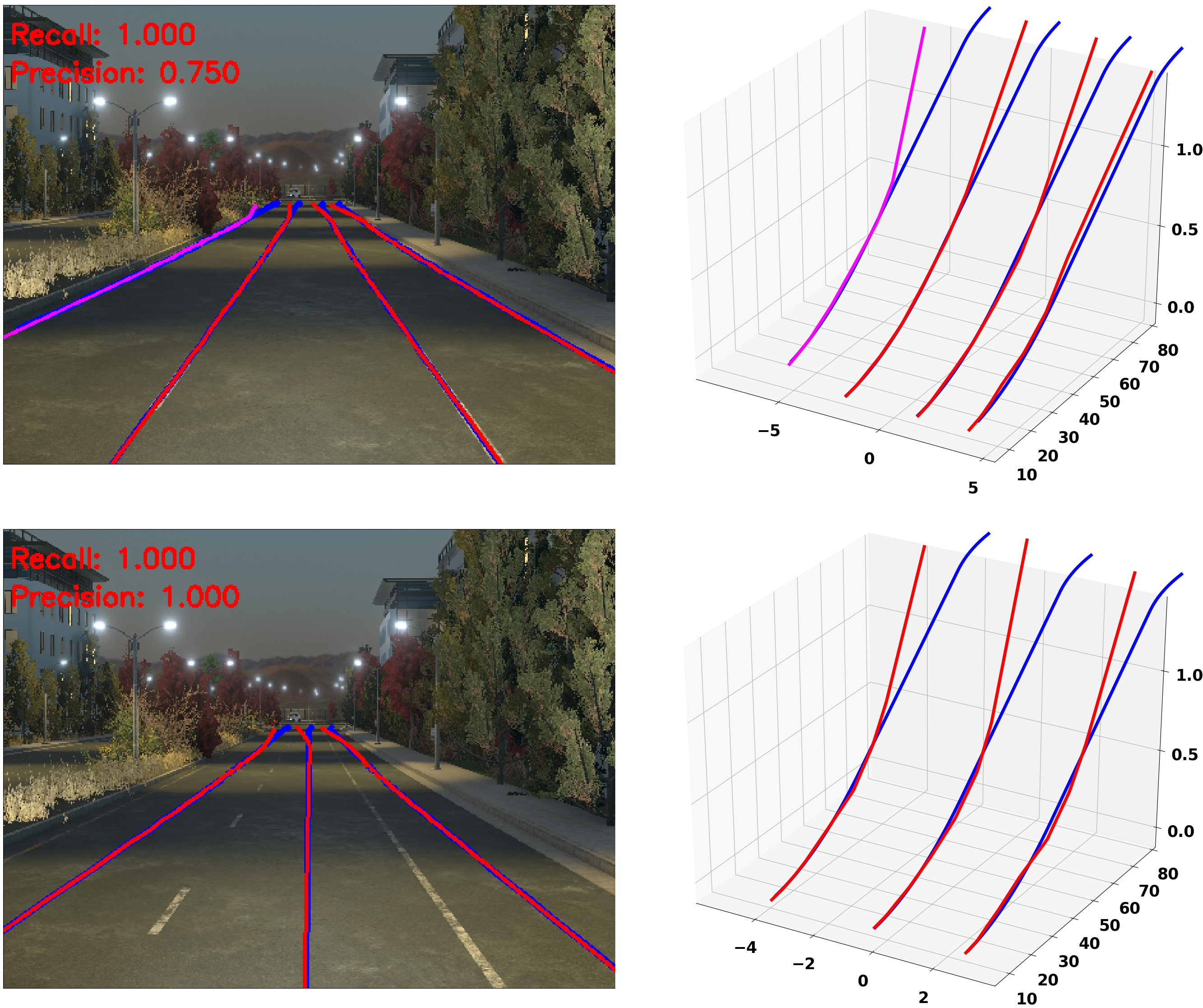}
  \HRule\\
  \includegraphics[width=0.33\textwidth]{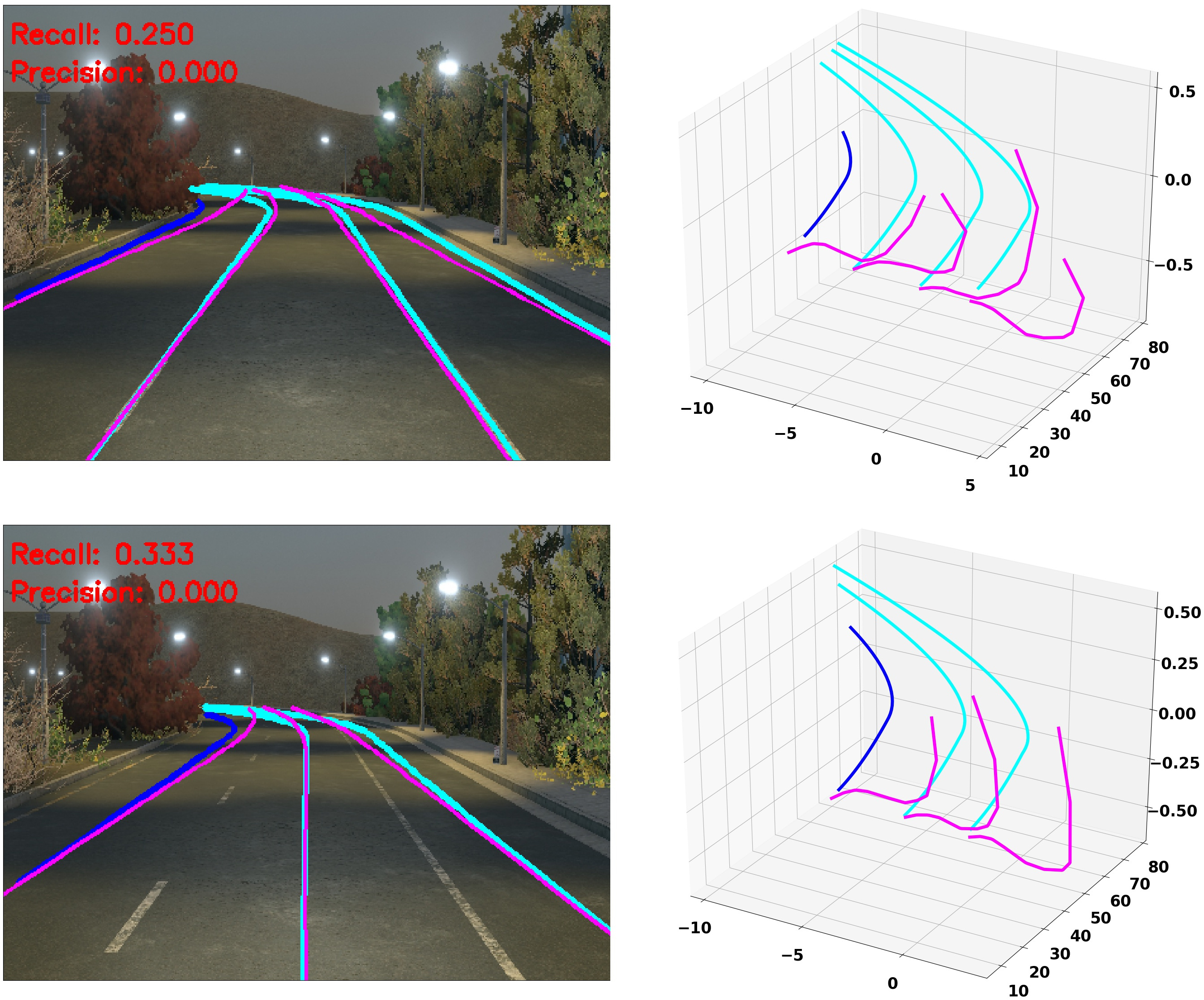} \hspace{1cm}
  \includegraphics[width=0.33\textwidth]{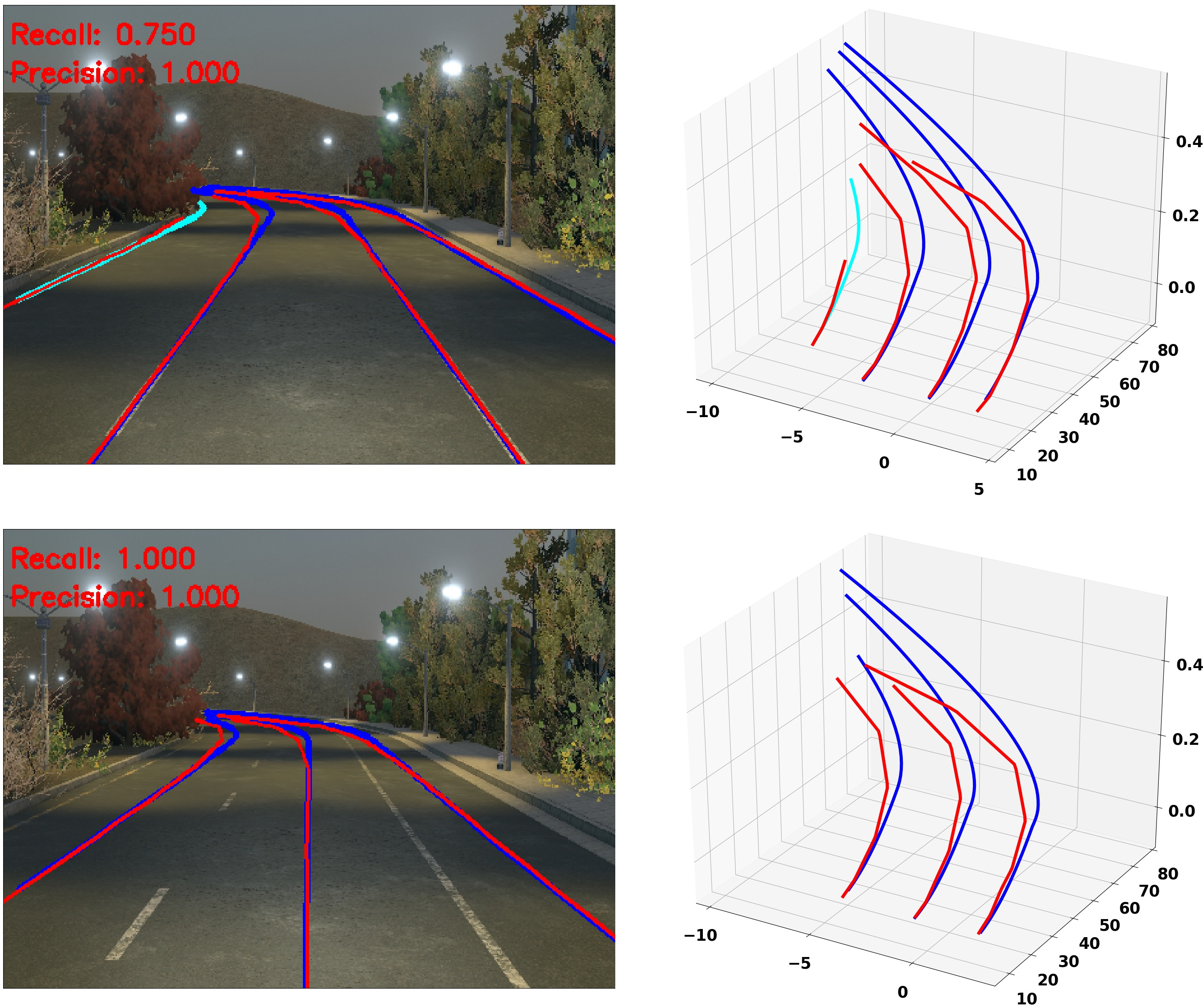}\\
  \hspace{10pt} 3D-LaneNet  \hspace{100pt} Gen-LaneNet
\caption{Visual comparison between {\bf 3D-LaneNet and Gen-LaneNet} are show on four examples. Examples are chosen from the data split evaluating an algorithm's robustness to illumination change. Observe that 3D-LaneNet is very sensitive to illumination change while Gen-LaneNet is not. For each example, we show image results on the left and 3D results on the right, lane line results in the top row, and center line results in the bottom row. We color the ground truth lanes in blue and the predicted lanes in red. While the purple lanes are missed detections and the cyan lanes are false alarms.}
  \label{fig:visual:3}
\end{figure*}

We provide qualitative comparison of both lane lines and center lines in two sets. First, we compare the original 3D-LaneNet and its improved version adopting our new anchor. This set of comparison is meant to emphasize the effect of our new anchor. The visualized examples are selected from the test set of the standard five-fold split of dataset. Observed from Figure~\ref{fig:visual:1}, the new anchor leads to consistency improvement over hilly and sharp-turning roads. Second, we present visual comparison of 3D-LaneNet and Gen-LaneNet as whole systems. The examples are chosen from the split of dataset considering seances with visual variation. As observed from Figure~\ref{fig:visual:3}, 3D-LaneNet can be very unstable encountering unobserved illumination, however Gen-LaneNet is rather robust.

\end{document}